\pdfoutput=1

\documentclass[11pt]{article}

\usepackage{acl}

\usepackage{times}
\usepackage{latexsym}
\usepackage{color,soul}

\usepackage{tcolorbox}
\newtcolorbox{mybox}[1]{colback=gray!1!white,colframe=gray!85!black,fonttitle=\bfseries,title=#1}

\usepackage[T1]{fontenc}

\usepackage[utf8]{inputenc}

\usepackage{microtype}

\usepackage{inconsolata}

\usepackage{csquotes}
\usepackage{booktabs}
\usepackage{multicol}
\usepackage{multirow}
\usepackage{graphicx}
\usepackage{subcaption}
\usepackage{amsmath}
\usepackage{cleveref}
\usepackage{tabularx}
\usepackage{amssymb}
\usepackage[inline]{enumitem}

\usepackage{todonotes}
\usepackage[normalem]{ulem}

\makeatletter
\newcommand*\iftodonotes{\if@todonotes@disabled\expandafter\@secondoftwo\else\expandafter\@firstoftwo\fi}  %
\makeatother

\newcommand*\samethanks[1][\value{footnote}]{\footnotemark[#1]}

\title{From Insights to Actions: \\ The Impact of Interpretability and Analysis Research on NLP}

\author{Marius Mosbach\textsuperscript{1,2}\,
 Vagrant Gautam\thanks{Authors contributed equally.}\textsuperscript{3}\,
 Tom\'{a}s Vergara-Browne\samethanks\textsuperscript{4,5}\\\textbf{Dietrich Klakow}\textsuperscript{\textbf{3}}\,~\textbf{Mor Geva}\textsuperscript{\textbf{6}} \ \vspace{3px} \\ \vspace{3px}
 \textsuperscript{1}Mila Quebec AI Institute~~~
 \textsuperscript{2}McGill University~~~
 \textsuperscript{3}Saarland University~~~\\
 \textsuperscript{4}Pontificia Universidad Cat\'{o}lica de Chile~~~
 \textsuperscript{5}CENIA~~~
\textsuperscript{6}Tel Aviv University\\
\small{{\tt  marius.mosbach@mila.quebec}\quad {\tt morgeva@tauex.tau.ac.il}}
}

\begin{document}

\maketitle

\begin{abstract}
Interpretability and analysis (IA) research is a growing subfield within NLP with the goal of developing a deeper understanding of the behavior or inner workings of NLP systems and methods.
Despite growing interest in the subfield, a criticism of this work is that it lacks actionable insights and therefore has little impact on NLP.
In this paper, we seek to quantify the impact of IA
research on the broader field of NLP.
We approach this with a mixed-methods analysis\footnote{Code and data publicly available at \url{https://github.com/mmarius/interpretability-impact}} of:
(1) a citation graph of 185K+ papers built from all papers published at ACL and EMNLP conferences from 2018 to 2023, and their references and citations, and (2) a survey of 138 members of the NLP community.
Our quantitative results show that IA work is well-cited outside of IA, and central in the NLP citation graph.
Through qualitative analysis of survey responses and manual annotation of 556 papers, we find that
NLP researchers build on findings from IA work and perceive it as important for progress in NLP, multiple subfields, and rely on its findings and terminology for their own work.
Many novel methods are proposed based on IA findings and highly influenced by them, but highly influential non-IA work cites IA findings without being driven by them.
We end by summarizing what is missing in IA work today and provide a call to action, to pave the way for a more impactful future of IA research.\looseness=-1
\end{abstract}

\section{Introduction}
\label{sec:introduction}

The rapid progress made in the development of large language models (LLMs, \citet{devlin-etal-2019-bert,radford2019language,raffel-etal-2020-t5,bommasani2022opportunities,touvron2023llama,openai2024gpt4,geminiteam2024gemini})
has had a profound impact on the field of natural language processing (NLP) \citep{gururaja-etal-2023-build}.
While these models demonstrate unprecedented performance and novel capabilities \citep{brown-etal-2020-language,wei2022emergent}, and are rapidly finding their way into real-world applications \citep{chatgpt,copilot,generative-search}, they are also largely treated as black boxes, which does not satisfy other expectations for successful machine learning deployment, such as trust, accountability, and explainability~\citep{lipton-2018-mythos,goodman-etal-2017-european}.
 
\begin{figure}[t]
    \centering
    \includegraphics[width=1.0\columnwidth]{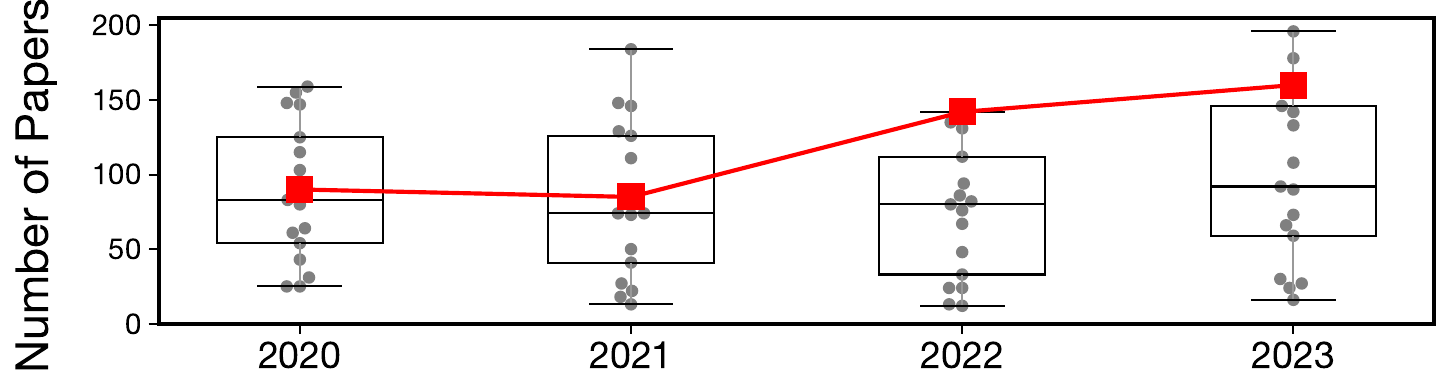}
    \includegraphics[width=1.0\columnwidth]{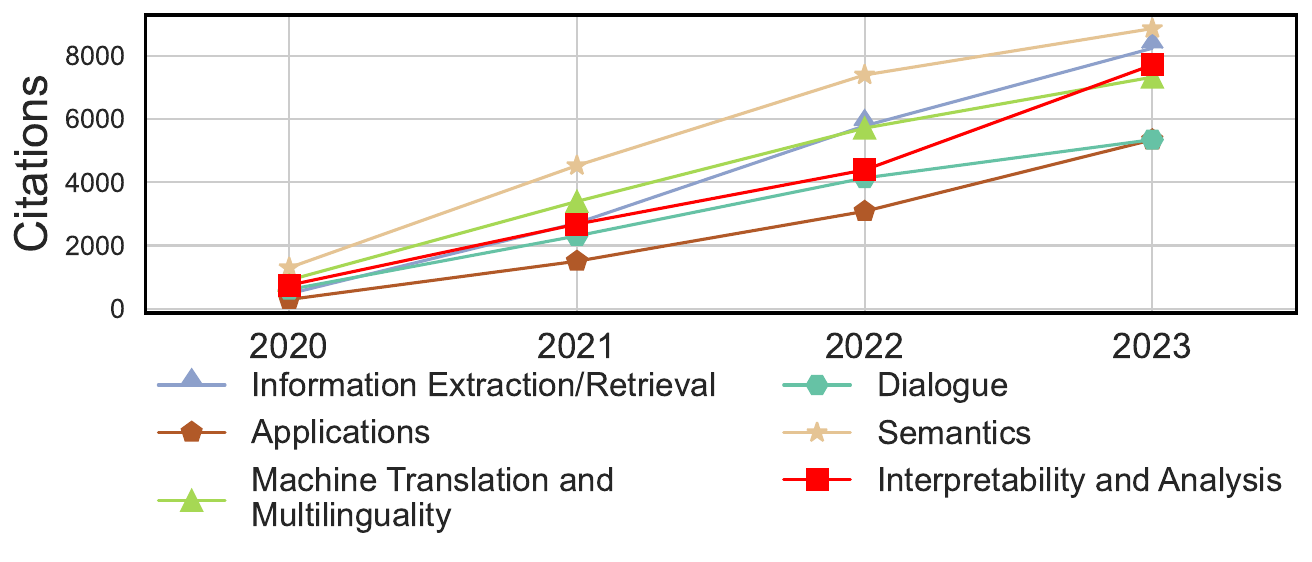}
        \caption{Interpretability and analysis (IA) is an increasingly popular subfield of NLP:
        (top) Number of IA papers in ACL/EMNLP in comparison to other tracks that have existed since 2020.
        The number of IA papers has grown considerably, from 90 papers in 2020 to 160 papers in 2023 (a growth rate of 77.8\%). This is the highest growth rate among these tracks. 
        (bottom) 
        Citations to IA papers compared to other highly cited tracks.\looseness-1
        }
        \vspace{-3mm}
    \label{fig:n-papers-per-track}
\end{figure}

In NLP research, these factors have motivated a large body of work on \textit{interpretability and analysis} (IA), which aims to understand the inner workings of LLMs and explain their predictions \citep[\textit{inter alia}]{belinkov-glass-2019-analysis,rogers-etal-2020-primer, räuker-etal-2023-toward}.
Many researchers in this area believe that better understanding LLMs is imperative to improve their efficiency, robustness, and trustworthiness, towards successful, safe deployment.
IA research has thus witnessed rapid growth in recent years and is now one of the biggest research areas 
(in terms of number of publications and citations) 
at major NLP conferences (see \Cref{fig:n-papers-per-track}).\looseness=-1

\noindent Despite the rapid growth of IA research (see also \Cref{fig:growth-per-track}),
a criticism of this work is that it often lacks actionable insights, especially for how to improve models, and therefore has little impact on how new NLP models are designed and built~\citep{räuker-etal-2023-toward,rai2024practicalreviewmechanisticinterpretability}.
This criticism raises questions about
whether its current form is the right path towards progress in NLP.

In this work, we tackle these questions with a systematic, mixed-methods study of the impact of IA research on NLP in the past and the present, and use our findings to inform a vision for the future of IA.
More specifically, we ask: \textbf{how does interpretability and analysis research influence NLP researchers in what they choose to work on, 
what they cite, 
and how they think about NLP altogether?}\looseness=-1

We perform a bibliometric analysis of 185,384 publications based on the two major NLP conferences, ACL and EMNLP, between 2018 and 2023, and solicit opinions from 138 members of the NLP community via a survey.
In addition to quantitative results, we perform qualitative analysis of survey responses and 556 papers.
This approach gives us a holistic view of the impact of IA research on NLP.

Our analysis reveals that \textbf{(1)} NLP researchers build on findings from IA work in their research, regardless of whether they work on IA themselves or not  (\S\ref{sec:benefits}),
\textbf{(2)} NLP researchers and practitioners perceive IA work to be important for progress in NLP, multiple subfields, and their own work, for various reasons (\S\ref{sec:importance}),
and \textbf{(3)} many novel non-IA methods are proposed based on IA findings and highly influenced by them, for various areas, even though highly influential non-IA work is not driven by IA findings despite citing them (\S\ref{sec:positive-examples}).\looseness-1

While our findings show that IA work presents insightful observations, there are still opportunities for greater impact on the rest of NLP.
Thus, based on survey responses, we identify the key ingredients that are missing in IA research today --- unification; actionable recommendations; human-centered, interdisciplinary work; and standardized, robust methods --- and close with a call to action and recommendations (\S\ref{sec:going-forward}).
We hope our work paves the way towards a more impactful future for IA research as the field continues to grow.

\section{Methodology}
\label{sec:scope}

We start by discussing what we consider as IA research and our approach for measuring impact.

\subsection{Interpretability and analysis (IA) research}

\textit{Interpretability} research has a long tradition in machine learning as well adjacent fields like NLP \citep[\textit{inter alia}]{tishby_zaslavsky_2015,karpathy2015visualizing,kim2018deep}.
There is no single agreed upon definition of the term \textit{interpretability} (see \citet{lipton-2018-mythos} and \citet{li-et-al-interpretable-deep-learning} for a critical discussion), but two prominent types of interpretability research focus on post-hoc explainability or increasing the transparency of machine learning methods and models \citep{lipton-2018-mythos,Madsen2024InterpretabilityNA}.
\textit{Analysis} research is an even broader term and one might argue that nearly every scientific paper contains some form of analysis.
In NLP, however, many interpretability and analysis papers have in common that their \textit{primary} contribution is an analysis that aims to advance our understanding of NLP in some way, e.g., by analyzing methods, models, or algorithms \citep{belinkov-glass-2019-analysis,rogers-etal-2020-primer}.\looseness=-1

Here, we adopt a broad definition of interpretability and analysis (IA) research in NLP that includes all papers that aim to \textbf{develop a deeper understanding} of the behavior or inner workings of NLP models, methods, or systems. 
This includes work on explaining models’ predictions or internal computations, investigating broader phenomena observed during pre-training or adaptation, and providing a better understanding of the limitations and robustness of existing models.

\subsection{Measuring impact}
\label{sec:impact}

Our goal is to measure the \textit{impact} of IA work on NLP research,
which is not trivial to define or quantify.
For a \textbf{holistic view of impact}, we consider two complementary ways of measuring it -- a bibliometric analysis, and a survey of the NLP community.

\paragraph{Citational impact}

In scientometrics research, citation counts are used as a standard measure of scientific impact \citep[\textit{inter alia}]{nicolaisen-2007-citation,bornmann-daniel-2008-citation,chacon-etal-2020-comparing}.
Thus, we perform a bibliometric analysis to quantify the citational impact of IA work on NLP research.\footnote{This choice excludes other forms of impact such as increasing user trust, influencing policy and regulation, etc. In addition, even though IA work impacts other fields, this is beyond the scope of our paper.}
We note that citation behavior is complex and there is a growing consensus that citation statistics might not be sufficient for measuring impact \citep{bornmann-daniel-2008-citation,zhu-etal-215-measuring,iqbal-etal-2021-a-decade}.

\paragraph{Surveying the NLP community}

\begin{figure*}[ht]
    \centering
    \includegraphics[width=0.95\textwidth]{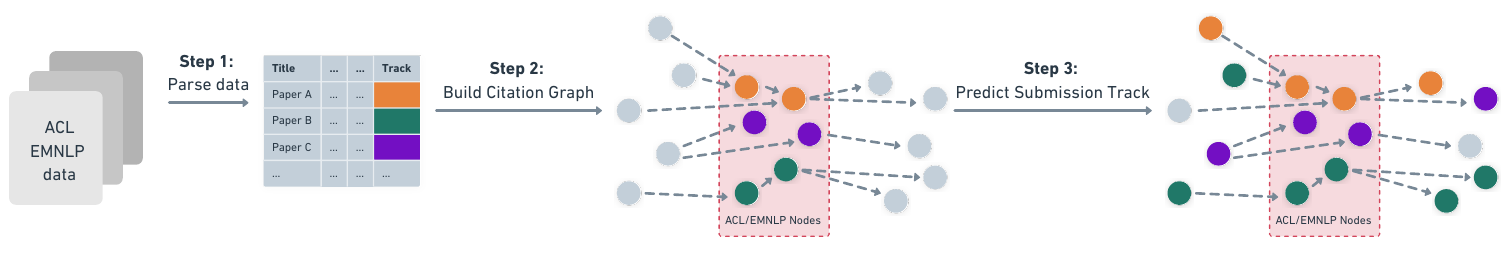}
        \caption{Diagram showing the process of constructing our citation graph. Starting from an initial set of ACL and EMNLP papers between 2018 and 2023, we collect their citations and references via the Semantic Scholar API and label the submission track of the papers with a classifier.}
    \label{fig:citation-graph-construction}
\end{figure*}

To incorporate a second dimension of impact beyond citation counts, we survey NLP researchers and practitioners on how they view the impact of IA research on the field.
Specifically, we ask respondents about their \textit{perceptions} of IA (its importance in general, for specific subfields, and its impact on progress in NLP), and their \textit{use} of IA (how much they read, are influenced by, and use concepts from IA work).
We also solicit opinions on what is missing in IA research and where it should go in the future.

\section{Citation graph and community survey}
\label{sec:methods}

Here, we describe the construction of our citation graph for bibliometric analysis, and the design of our survey of the community.

\subsection{Citation graph construction}
\label{subsec:citation}

As \Cref{fig:citation-graph-construction} illustrates, we begin constructing our citation graph from an initial set of all papers published at ACL and EMNLP from 2018 to 2023.
We focus on these two venues as they are leading NLP conferences with a dedicated track for interpretability and analysis research since 2020.\footnote{We discuss this decision in more detail in \Cref{sec:limitations}.}
We then use the Semantic Scholar API \cite{Kinney2023TheSS} to get all citations and references of these initial papers, and add them to our citation graph.
For papers outside our initial set (where we have gold labels), we use a classifier to predict their submission track.
More details on all these stages are provided below.

\paragraph{Collecting ACL and EMNLP papers}

We collect paper lists and track information from various sources (see Table \ref{tab:conference-papers-sources} in \Cref{sec:appendix:citation-graph}), as there is no one source of this data for ACL and EMNLP conferences.
Between 2018 and 2023, official names of submission tracks have changed substantially, so
we standardize all data to 27 tracks.
More details on this process are provided in \Cref{sec:appendix:citation-graph}, including summary statistics per track (\Cref{tab:research-track-distribution}).

\paragraph{Building the citation graph}

We collect the citations and references of each paper in our initial set via the Semantic Scholar API \citep{Kinney2023TheSS},
resulting in a citation graph of 185,384 papers (see \Cref{tab:citation-graph-statistics} in \Cref{sec:appendix:citation-graph} for additional statistics).
For each node (paper) in the graph, we store its title, abstract, and venue.
For each edge (citation), we store information on the citation intent (binary labels for background, use of methods or comparing results), and citation influence (normal vs. highly influential), all of which are provided by Semantic Scholar.\looseness=-1

\paragraph{Labeling the citation graph}

To assign all papers in the citation graph to our standardized set of tracks,
we train a classifier based on the titles and abstracts from our initial set of papers. We find that some tracks are very hard to predict due to limited training data and the inherent ambiguity of submission tracks. We thus keep 11 well-performing labels (including IA), and introduce an `Other' label to group the remaining papers.

Our final classifier achieves a test micro/macro-F1 score of 0.61/0.61.
Although this appears low, we note that submission tracks have fuzzy boundaries and papers can often be plausible submissions to multiple tracks.
Given that we care primarily about predicting IA compared to other tracks, we evaluate our classifier on two additional sets of gold data, and obtain 78.1\% and 87.8\% accuracy on each set.
We provide further details on classifier construction and evaluation, and we verify our findings with only gold labeled papers in \Cref{sec:appendix:citation-graph}.\looseness=-1

\subsection{Surveying the NLP community}
\label{subsec:questionnaire}

To solicit opinions from the NLP community on the impact of IA research, we ran a survey from March 19th to June 7th, 2024, advertising within our networks, on social media, and on NLP mailing lists.
The full survey is shown in Appendix \ref{sec:appendix:survey}.

\begin{figure*}[t]
    \centering
    \includegraphics[width=0.95\textwidth]{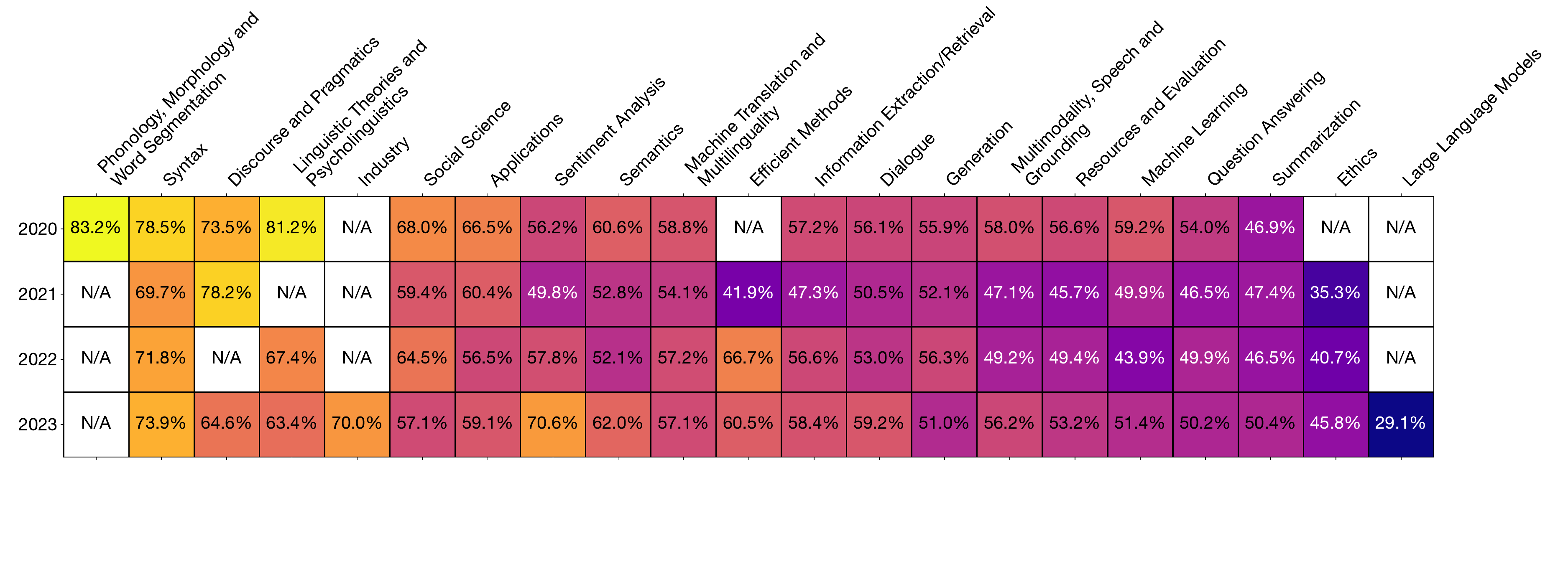}
        \vspace{-10mm}
        \caption{
        CSI scores for the interpretability and analysis track are favorable (> 50\%) when compared to other tracks.
        The CSI score represents the probability that a random interpretability and analysis paper published in certain year has more citations than a random paper of other track published the same year.}
    \label{fig:csi}
    \vspace{-3mm}
\end{figure*}

To strike a balance between easy scoring and respondent expressivity, we included multiple-choice as well as optional free response questions~\citep{Shaughnessy_Zechmeister_Zechmeister_2015}.
We refined the survey following best practices\footnote{We made sure to clarify definitions, avoid leading questions, etc.~\citep{Shaughnessy_Zechmeister_Zechmeister_2015}.} and with feedback from four senior NLP researchers who filled out a pilot version.
We received a total of 138 responses from NLP researchers in academia and practitioners in industry, with 61\% of respondents not working on IA themselves (see \Cref{sec:appendix:survey} for more statistics).

Two authors performed qualitative coding, an inductive method from the social sciences~\citep{Saldana2021-ki}, to identify themes in answers to the free-response questions.
More details on the coding process are provided in \Cref{sec:qualitative-coding}.
We measure inter-coder reliability with percentage agreement~\citep{oconnorjoffe2020icr}, which was above 90\% across all subsets of annotation.

\section{Researchers build on findings from IA research in their work}
\label{sec:benefits}

We begin by analyzing whether researchers \textit{use} contributions of IA research in their work. We approach this by analyzing citational use, as well as survey-reported use beyond citations.

\paragraph{IA papers are cited more often than other tracks}

When comparing papers from different tracks, global counts of citations can be misleading, as a small number of papers can account for most of the citations in a field \cite{ioannidis2016citation}.
To account for this, we compare citations based on the \textit{Citation Success Index} (CSI; \citealp{milojevic-etal-2017-citation-success}) metric.
Given two groups of papers $A$ and $B$, the CSI score computes the probability that a random paper from $A$ is more cited than a random paper of $B$.
This score is not subject to biases from the skewness of the citation distribution, and it is clearly interpretable;
e.g., if we draw random IA and Machine Translation papers from EMNLP or ACL in 2023,
there is a 57.1\% chance that the IA paper is more cited than the Machine Translation paper.\looseness-1

\Cref{fig:csi} shows that CSI scores for the IA track are often favorable ($\text{CSI} > 50\%$) when compared to other tracks.
In 2023, only the Ethics and the Large Language Models tracks had favorable CSI scores when compared to IA.
This shows that IA papers have higher citational impact than other tracks.

\paragraph{IA papers are well cited outside of IA}

\begin{figure}[t]
    \centering
        \includegraphics[width=\columnwidth]{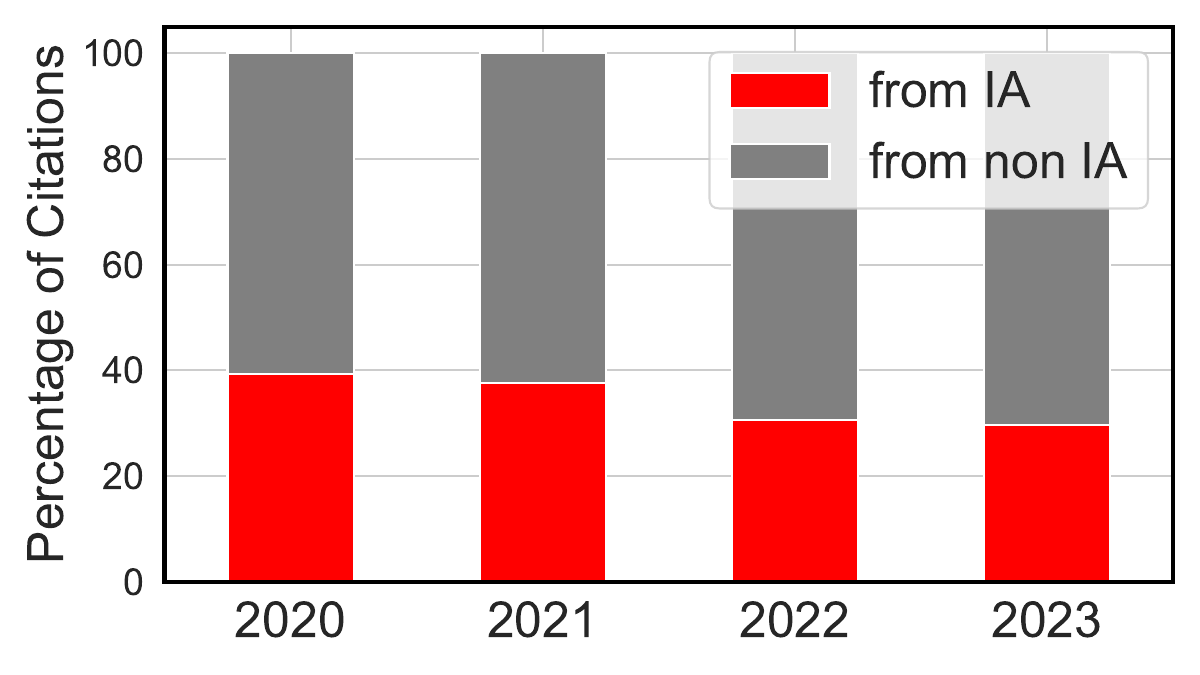}
        \vspace{-7mm} \caption{Origin of citations to IA papers in our citation graph.
        More citations come from non-IA work than IA work, showing citational impact beyond the subfield.\looseness=-1
        }
    \label{fig:interp-vs-non-interp}
    \vspace{-3mm}
\end{figure}

While high CSI scores tell us that IA papers are cited well, they do not tell us where these citations are coming from, i.e., are IA papers mostly cited by other IA papers or by papers outside of IA? To evaluate the impact of IA work outside of IA, we compare citations within the same track, which we call \textit{intra-track} citations, to \textit{extra-track} citations, i.e., citations from outside the track.

\Cref{fig:interp-vs-non-interp} shows that most citations to IA papers are predicted to be extra-track citations.
The proportion of references to IA papers differs considerably by citing track, with papers about Efficient Methods, Machine Learning, and Large Language Models citing IA research more frequently than others (see \Cref{fig:tracks-citing-interp}).
While the IA track does not stand out in terms of its \textit{extra-track} citations compared to other tracks (see \Cref{fig:intra-track}), these results still demonstrate that the citational impact of IA research extends well beyond the IA track itself.

\paragraph{IA papers are central in NLP}

Next, we assess whether IA papers are impacting NLP as a whole rather than just specific tracks.
We quantify this with
the \textit{Betweenness Centrality} (BC) metric, a measure of \textit{interdisciplinarity} \cite{leydesdorff2007betweenness, barnett2011citations, leydesdorff2018betweenness}.
BC quantifies the extent to which a node in the graph acts as a \textit{bridge} along the shortest path between two other nodes \cite{golbeck2015introduction};
nodes with higher BC are considered more important as more information passes through them.\footnote{We provide further discussion of BC in \Cref{sec:normalizations}.}
Therefore, we interpret papers with a high BC as \textit{important} papers that are essential for the connectivity of the citation network.\looseness=-1

We compute the BC for every paper in EMNLP and ACL since the IA track started (2020), and find that the median BC of IA papers is higher than most other tracks, at $1.23 \times 10 ^ {-7}$.
Notably, IA ranks as the second most central track overall, following the Large Language Models track, which has a median BC of $1.95 \times 10 ^ {-7}$.
These results (shown in \Cref{fig:betweenness-centrality}) provide further evidence that IA work plays a central role in the ACL/EMNLP citation network.\looseness-1

\paragraph{IA influences the work of NLP researchers}

For a complementary view of impact beyond citations, we survey NLP community members on how often they use concepts from IA in their day-to-day work, and more broadly, how IA influences their work.

\begin{figure}[t]
    \centering
    \hspace{-5mm}
    \includegraphics[width=\columnwidth]{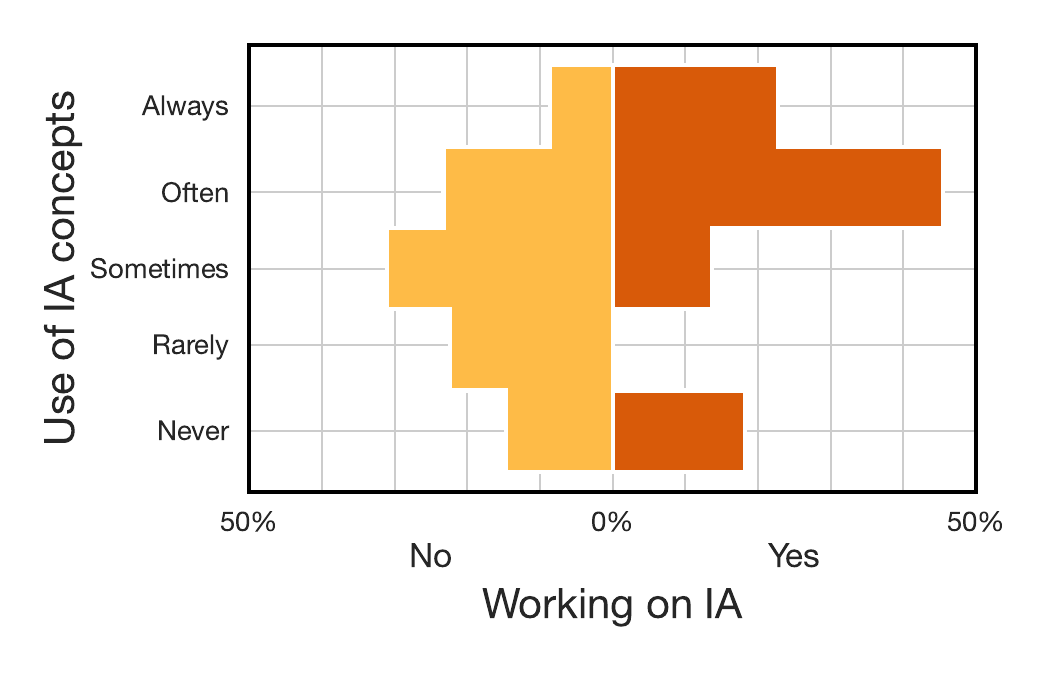}
    \caption{Survey responses on the frequency of using concepts from IA research, split by whether the respondents work in this field or not.}
    \label{fig:concepts-work}
\end{figure}

As \Cref{fig:concepts-work} shows, the median rating for use of IA concepts by respondents who work on IA is \textit{often}, while even the median respondent who doesn't work on IA uses concepts from IA \textit{sometimes}.
In both groups of respondents, there are people who \textit{always} use IA concepts in their day-to-day work.
Beyond this, IA work influences respondents in different ways:
it provides respondents with research ideas (91\% of respondents who work on IA; 60\% of respondents who don't), changes mental models of model capabilities and limitations (77\%; 65\%), and helps ground explanations of respondents' results (64\%; 59\%).
Notably, only 9 (6.5\%) respondents state that IA does not affect their work.
These results complement our citation-based findings by providing further evidence that IA work impacts both IA and non-IA researchers and their research.

\section{Researchers find IA work important}
\label{sec:importance}

We continue by surveying the \textit{perceived importance} of IA work by the NLP community.
We consider various perspectives, such as the perceived importance of IA research on overall progress in NLP as well as on individual subfields.
133 out of 138 respondents consider \textbf{IA work important}, and perceive it as important \textbf{for progress} in NLP, \textbf{multiple subfields}, and \textbf{for various reasons}.

\paragraph{Perceived importance for progress in NLP}

\begin{figure}[t]
    \centering
    \includegraphics[width=\columnwidth]{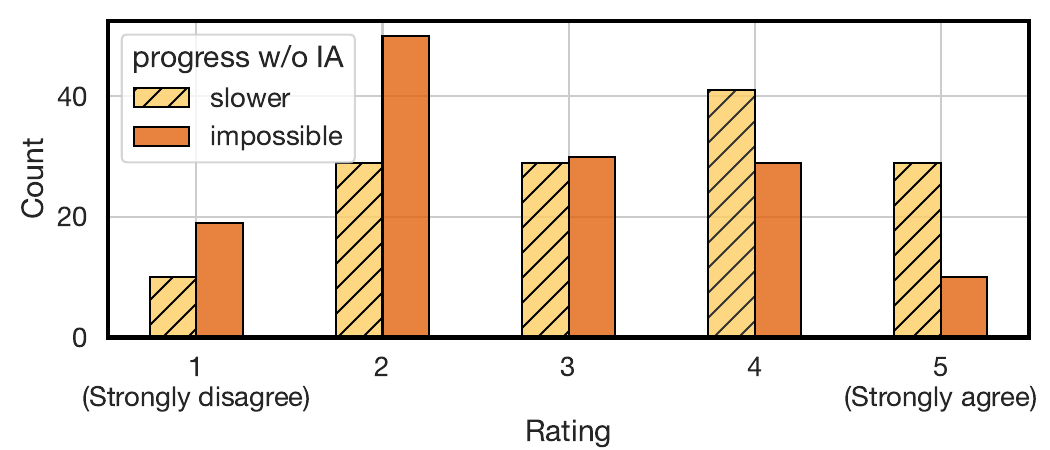}
    \caption{Survey responses (N=$138$) on whether progress in NLP in the last 5 years would have been \textit{slower} or \textit{impossible} without findings from interpretability and analysis research.
    Respondents believe progress in NLP would be slower but not impossible without IA.
    }
    \label{fig:slower-impossible}
\end{figure}

Figure \ref{fig:slower-impossible} shows that most respondents agree that without IA findings, progress in NLP in the last 5 years (2019 to 2024) would have been slower, but not impossible. Surprisingly, it appears that \textit{people who are more deeply engaged with interpretability are more critical of it}.
Respondents who read
more IA work than other topics in NLP, respondents who often or always use concepts from IA literature, and respondents who work on IA themselves all rate IA as having a lower impact on progress in NLP than those who read less IA, use related concepts less frequently, and who work on other topics.

It is plausible that respondents who are more engaged with IA work know it better and thus give better-calibrated impressions of the field as a whole, which happen to be more critical.
However, it is worth noting that they are perhaps forming their opinions from a different sample of papers (i.e., the average paper from a large body of work) than those who are less engaged with IA work, whose reading might be skewed towards IA work that is more highly cited and influential.
This also raises the question of how IA or indeed any subfield \textit{should} be evaluated -- by the average paper in it, or by the ones that stand out?

\noindent There are many other factors that could also influence the results we see, e.g., that respondents in different categories are reading IA papers that deal with different topics, that they have different levels of research experience, and that they have different definitions of ``progress'' in NLP.
See \S\ref{sec:limitations} for a discussion of these factors.

\paragraph{Perceived importance for different subfields}

\begin{figure}[t]
    \centering
    \includegraphics[width=\linewidth]{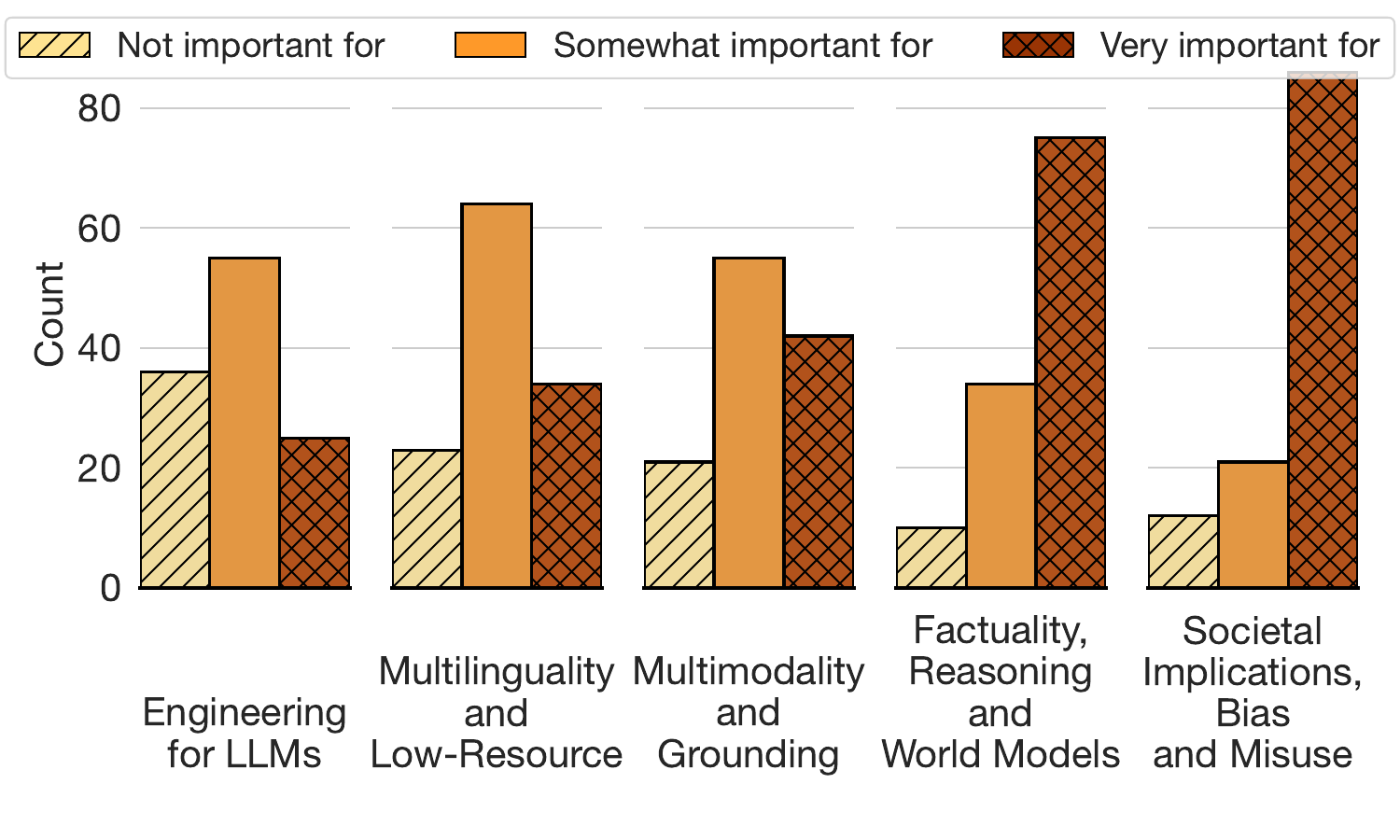}
    \caption{Survey responses (N=138) on how important interpretability and analysis research is to work in different subfields.
    IA is considered most important for work on reasoning, factuality, and bias, and least imporatnt for LLM engineering.
    }
    \label{fig:subfields-importance}
\end{figure}

\Cref{fig:subfields-importance} shows that the IA work is perceived as being important to differing extents for other subfields within NLP.
The modal response is that IA work is \textit{somewhat} important for work on multilinguality (52\% of responses), multimodal learning (47\%) and engineering for large language models (47\%), and that it is \textit{very} important for work on reasoning (63\%) and bias (72\%).
Of the five subfields we consider, engineering for LLMs is perceived to be least impacted by IA work, with 31\% of respondents indicating that they think IA work is not important for it.
These findings are consistent with the themes we find in papers that are highly influenced by IA research, where bias and reasoning are well-represented, and pre-training and architectural advancements appear less frequently.

\paragraph{Reasons for importance}

When asked whether they thought IA work was important and if so, why, respondents overwhelmingly (133/138) consider it important, citing a variety of reasons, the most popular of which were: understanding model limitations and capabilities (90\% of respondents), explainability for users (66\%), improving model trustworthiness (59\%), and improving model capabilities (50\%).
While a small percentage (4.3\%) of respondents indicated that they thought it was not important (possibly also due to selection bias in our survey), we found that they
voice the same concerns as those who do find it important,
e.g., a lack of actionability, results that do not scale, and a lack of impact on the most capable models of today.
In our recommendations for the future of the field (\S\ref{sec:going-forward}), we go into these in more detail.

\section{A closer look at influential papers}
\label{sec:positive-examples}

So far we have discussed findings about IA as a whole, either by considering the role of IA papers in the ACL/EMNLP citation graph or the perception of IA work within the community.
In this section, we 
zoom in on specific influential papers sourced from both our survey and citation graph. We seek to answer: What are these papers about? What kind of work are they impacting, and how?

To this end, we inductively obtain the themes of a total of 585 papers, through qualitative coding of their titles and abstracts by two authors~\citep{Saldana2021-ki}.
The 585 papers include:
\textbf{(1)} All papers mentioned more than once as having influenced survey respondents' work (N=29); \textbf{(2)} highly-cited IA papers from our citation graph (N=50); \textbf{(3)} highly-cited non-IA papers from our citation graph (N=50); \textbf{(4)} non-IA papers that cite and are highly influenced by the top-10 most-cited IA papers (N=456).
The resulting themes are mostly descriptive, including topics (e.g., \textit{in-context learning}, \textit{training dynamics}) and contribution types (e.g., \textit{novel method}, \textit{analysis}).
Percentage agreement on our coded themes is above 90\% for each subset of papers. See \Cref{sec:qualitative-coding} for more details.

Our analysis reveals that beyond background citations, IA work influences the development of many novel models and metrics outside of IA work, and affects work in domains such as question answering (QA), reasoning, and bias.

\paragraph{What are influential IA papers about?}

Of the papers that survey respondents submitted as examples of work that has directly influenced their own work, representation analysis appears in over a third of the papers, novel methods for interpretability (e.g., causality, interventions, steering, neuron/activation analysis, etc.) are proposed in nearly a quarter of them, and probing also appears in 24\% of these papers.\looseness-1

In contrast, the top-50 most cited IA papers are more often about the \textit{analysis} component of IA (40\%).
Novel methods (for analysis, evaluation, linguistics, probing) are proposed in 26\% of papers, and evaluation is a main contribution of 32\%.
As expected, the most cited non-IA papers in our citation graph mostly consist of highly influential datasets, models, and methods, e.g., HotpotQA, BART, prefix-tuning~\citep{yang-etal-2018-hotpotqa,lewis-etal-2020-bart,li-liang-2021-prefix}.
More top themes are shown with the percentage of papers in \Cref{tab:influential-paper-themes} in \Cref{sec:appendix:additional-results}.\looseness-1

We also find evidence that many IA papers create novel metaphors to understand models --- e.g., seeing feed-forward layers as key-value memories \citep{geva-etal-2021-transformer}, or reading from and writing to the ``residual stream''~\citep{elhage2021mathematical}, and many analysis papers highlight the limits of models.
As survey respondents cited these very reasons for why they perceive IA work as important, these themes corroborate why these papers would be particularly influential.
In addition, many of the qualities that survey respondents feel are currently lacking in IA research (see \S\ref{sec:going-forward}) appear in these papers, such as moving beyond toy models~\citep{wang2023interpretability}, and providing actionable methods~\citep{meng2022locating}.\looseness=-1

\paragraph{Why are influential IA papers cited?}

As citations can have a variety of reasons \citep{zhu-etal-215-measuring,tahamtan-bornmann-2019-what}, we examine three types of citational intent -- background, methods and results citations (see \Cref{fig:citation-intent-top-papers} in \Cref{sec:appendix:additional-results}).
Overall, we find that influential IA papers are cited most often as background citations, then as methods citations, and least frequently when comparing results.
In comparison, highly cited papers that are \textit{not} about IA tend to be cited most frequently for methods.
This is expected, as many of these papers are about popular datasets and models, as described above.\looseness-1

\paragraph{What are the citing papers about?}

Despite the large number of background citations, however, there is plenty of work---including non-IA work---that is highly influenced (according to Semantic Scholar) by IA research.
For a closer look at what these citing papers do, we analyze all 456 papers with highly influential citations to one of the top 10 most-cited IA papers, and annotate their themes based on titles and abstracts.

Unsurprisingly, many of the papers have themes in common with what they cite, e.g., papers that analyze multilingual models are frequently cited by papers on cross-lingual transfer.
We thus focus on the \textit{difference} in themes between citing papers and cited papers, and find that \textbf{over 33\% of non-IA papers that are highly influenced by IA work propose novel methods}, e.g., many novel ICL methods cite analysis work on demonstrations \cite{min-etal-2022-rethinking} and similarly, many novel methods for bias mitigation cite datasets for stereotype evaluation such as \citet{nangia-etal-2020-crows} and \citet{nadeem-etal-2021-stereoset}.
These provide concrete counterexamples to the claim that IA work does not influence modeling improvements.

\paragraph{Is IA work impacting highly cited non-IA work?}

Looking at the highly-cited non-IA papers, we find that these too tend to cite IA work frequently.
22 out of the top 50 most cited non-IA papers are even highly influenced by some IA work, but 28 are not highly influenced by \textit{any} IA work.
These results show that while highly influential non-IA work does acknowledge IA findings, it is likely not driven by them.\looseness-1

\section{Main takeaways and discussion}
\label{sec:going-forward}

We end by discussing our main findings and recommendations on how to move IA research forward.

\paragraph{Main takeaways}

In \S\ref{sec:benefits}, we saw that \emph{IA research plays a central role in NLP} and researchers build on findings from IA work in their research, regardless of whether or not they work on IA themselves.
In \S\ref{sec:importance}, we saw that \emph{NLP researchers and practitioners perceive IA work to be important} for progress in NLP, and multiple subfields.
They also find it important for their own work for a variety of reasons, regardless of whether they work on IA themselves. Finally, we took a closer look at the most influential IA papers in \S\ref{sec:positive-examples} and found that \emph{many novel methods are proposed based on IA findings and highly influenced by them}, for various areas---in particular, work on reasoning, factual knowledge, and bias.
All these findings present a very positive view of IA research and its role within NLP in the past and the present.
In the remainder of this section, we turn to the future of IA research.

\paragraph{What is missing?}

To understand what the NLP community believes to be important for the future of IA work, we asked survey respondents what they feel is missing in current IA work and what should be different going forward.
25\% of the responses to this question mentioned a lack of big picture and unified understanding in IA work. For example, one respondent said:
\begin{displayquote}
\textit{``I think the focus should be on climbing the right hill towards a higher level understanding instead of focusing on interesting individual behaviors.''}
\end{displayquote}
The next three most frequent concerns are a lack of utility (i.e., not being useful in practice),
modeling improvements and actionability---concerns that are also echoed by the respondents who do not find IA research useful for their own work.
Interestingly, a commonly voiced opinion among these participants is that they believe that scale and performance are all that is needed for good NLP models, and that IA work only has importance for understanding models rather than for building them.
Additionally, respondents mention that IA work could use more interdisciplinary connections, through collaboration with domain experts, user studies, and human-centered approaches to computing.

Finally, we note another theme appearing in 10\% of responses: as IA has a lack of consensus on reliable and trustworthy methods, it is unclear how such work should be evaluated.
Although this is not a new concern~\citep{belinkov-glass-2019-analysis}, it remains relevant for the impact of IA on NLP.

\paragraph{A call for action}

Based on our findings, we make the following recommendations for IA work:

\begin{mybox}{{Going forward, IA researchers should:}}
\begin{enumerate}
[itemsep=1pt, topsep=2pt, leftmargin=2.5pt]
    \item Think more about the big picture
    \item Strive for more actionable work
    \item Center humans in their research
    \item Work towards standardized, robust methods
\end{enumerate}
\end{mybox}

Concretely, big-picture thinking (1) involves working towards general truths about model architectures or behaviors, rather than model-specific results.
Future work should try to synthesize existing strands of research to unify their findings and viewpoints.
An example of what this might look like outside IA research is \citet{he2022towards}.\looseness=-1

Actionable work (2) requires thinking about how an IA finding can propel new ways of building/using NLP systems, rather than merely being descriptive.
More specific examples of this include research that uses interpretability findings to, e.g., improve the fairness of NLP systems, or make NLP models more efficient and robust.

Centering humans (3) entails evaluation with realistic, relevant data and tasks, and performing user studies and human evaluation.
Human-centered IA work can also be enhanced through interdisciplinary reading and collaboration.
An example for research that falls under this category is \citet{ivanova2024elementsworldknowledgeewok}, which proposes a cognition-inspired framework for evaluating LLM world knowledge.

Finally, we urgently need to build consensus on using and evaluating IA methods (4).
Rigorous, well-motivated methods (e.g., using causality) are critical, rather than correlative evidence that may not be correct or faithful.
We believe that standardized, robust and widely accepted methods will increase trust in IA work, and lead to the easier and wider adoption of IA methods.

Due to the constraints of space and time, we note that it would be difficult for one work to address all these points while still making a focused contribution.
Thus, we stress that our call to action is for IA research as a whole to revisit its priorities,  rather than a checklist for individual papers to address.

\paragraph{IA for its own sake}

In closing, we would like to highlight a viewpoint that came up multiple times in survey responses, which was to question the premise of this paper, i.e., to measure the impact of IA on NLP.
Many respondents noted that they see IA work as being a valuable scientific pursuit in its own right, stating that \textit{``Without it, we're not doing science,''} or \textit{``It's cool! That's enough for me.''} Respondents further criticized the often performance-focused definitions of utility, progress, and impact.
One respondent noted that these definitions of utility have been determined \textit{``by extrinsic sociological factors in the broader field of AI''}. We sympathize with this observation and note that the focus on performance is a feature of NLP at this point in time. What we value might change going forward, especially as NLP systems are increasingly part of our daily lives, and qualities such as robustness and fairness become even more important.

\section{Related work}
\label{sec:related-work}

The increasing number of IA publications during the last few years has resulted in several survey or position papers that critically discuss existing work, identify common patterns, and provide suggestions for how to go forward. \citet{lipton-2018-mythos} critically questions common motivations behind interpretability and the lack of definitions in the field. Following their recommendation, we provide a definition of what we consider interpretability and analysis research in \S\ref{sec:scope}. \citet{belinkov-glass-2019-analysis} summarize trends in early IA work and discuss recommendations for how to overcome the limitations of IA research. Similar to our work, they recommend that future work should think about better ways to evaluate IA research and findings. \citet{rogers-etal-2020-primer} survey and synthesize IA work on BERTology, a subfield of IA work that focuses on encoder-only language models. \citet{räuker-etal-2023-toward} survey a large number of papers that study the internals of language models (transparency), and discuss key challenges in the field. Like us, they also argue for better ways of evaluating IA methods, as well as more actionability  and grounding in real-world applications.
More recently, \citet{Madsen2024InterpretabilityNA} discuss two prominent trends in interpretability research (post-hoc explanations and intrinsic interpretability) and argue that interpretability (``the study of explaining models in understandable terms to humans'') needs a new paradigm centered around faithfulness.\looseness=-1

Several other works study citational patterns and trends within the broader NLP community. \citet{mohammad-2020-examining} uses citations to measure the impact of NLP publications indexed by the ACL Anthology. Like us, they compare how well papers from different areas within NLP are cited, and use citation statistics to draw conclusions about the impact of different subfields within NLP.
\citet{singh-etal-2023-forgotten} use citations as an indicator for how widely the community is reading. They demonstrate a recency bias in citation behavior with a study of temporal citation trends, i.e., a majority of cited papers fall within a five year time period before publication of the citing work. 
\citet{wahle2023we} analyze the influence between NLP and other fields over the years. Also using Semantic Scholar, they rely on citations to conclude that NLP has become more \textit{insular} over time.
Similarly, \citet{subramonian2024understandingdemocratizationnlpml} find low levels of extra-disciplinary citation when analyzing how NLP and ML researchers discuss democracy.
More specific to IA, \citet{jacovi-2023-trends} uses Semantic Scholar to curate a large number of papers focusing on explainability, studying citation trends in the field based on this collection.

Another set of related papers surveys the NLP community for their perceptions and opinions, a method we also use. \citet{gururaja-etal-2023-build}, for example, focus on paradigm shifts and study factors that shape NLP as a field. They conduct interviews with NLP researchers and experts and gather their opinions on critical trends and patterns that emerge in the field. \citet{pramanick-etal-2023-diachronic} also focus on paradigm shifts and impact, but from a diachronic perspective. They provide a novel framework to study the evolution of research topics within a field to establish what drives research in NLP across time, and they find that tasks and methods have a bigger impact on the field than metrics do.\looseness-1

Lastly, there are several papers in the scientometrics literature that study and compare the impact of research using the same metrics as we do: \citet{chacon-etal-2020-comparing} apply the citation success index to compare sub-fields in physics, and
\citet{leydesdorff2007betweenness} propose the use of Betweenness Centrality as a measure of the interdisciplinarity of journals.

\section{Conclusion}
\label{sec:conclusion}

We contribute a mixed-methods analysis of the impact of interpretability and analysis research on NLP.
By analyzing a citation graph of 185K+ papers built from all papers published at ACL and EMNLP from 2018 to 2023, surveying 138 respondents from the NLP community, and manually annotating 556 papers,
we found that IA work is well-cited in other subfields of NLP, central to the NLP citation graph, and highly influential to many novel methods.
NLP researchers and practitioners perceive IA work as important for progress in NLP, multiple subfields (especially reasoning and fairness), and for their own work.
In sum, even though highly influential models, methods and datasets are not driven by IA findings,
IA work still has a great impact on NLP in the past and the present.
We conclude with a call to action based on what is missing in the subfield, to pave the way for IA work to be even more impactful in the future.

\section*{Limitations}
\label{sec:limitations}

\paragraph*{Focus on papers published at ACL and EMNLP}

Although ACL and EMNLP are the most cited *CL venues~\citep{mohammad-2020-examining}, our analysis excludes several other big NLP venues, including EACL, NAACL, AACL, TACL, and BlackboxNLP, a workshop which focuses on IA work.
Additionally, given the growing interest in NLP, and in particular, LLMs, from the broader machine learning community, there is an increasing number of IA papers published at machine learning conferences such as ICLR, NeurIPS, and ICML, which we also do not consider in our analyses.
Similarly, a vast amount of work on mechanistic interpretability has been published as articles (e.g., on LessWrong\footnote{\url{https://www.lesswrong.com/}} and the AI Alignment Forum\footnote{\url{https://www.alignmentforum.org/}}), and blog posts (e.g., by Anthropic\footnote{\url{https://www.anthropic.com/}}). Therefore, there is a risk that our analysis misses potentially influential IA work published at these venues. 

This is mitigated to an extent by our survey, where respondents mention some of these papers and blog posts, which we then discuss in our paper.
In addition, the set of papers we consider for our analysis is very large (our initial set contains 477 IA papers). This makes us confident that the findings we draw from these papers (and those citing them) are representative of broader trends in the impact of IA research in NLP. We leave it to future work to investigate the impact of IA work published outside of established NLP venues.

\paragraph*{Focus on 2018 to 2024}

As our analysis focuses on papers published between 2018 and 2024,
our results represent a snapshot in time on the scale of research in NLP, where models and methods come and go.
The time period that we look at is dominated by transformer-based language models, and a paradigm of using large, general-purpose pre-trained models for many tasks, and thus many IA papers focus on studying these.
Understanding this as the context of our analysis and results is important, as they may look completely different in a time period where the most popular models and IA methods are different.
This also means that our results cannot speak to the impact of \textit{today's} IA work, which will only become clear in the future.

\paragraph*{Not all citations are equal}

Although our use of citations is an important component of how we quantify impact in this paper, we do not consider citational context or distinguish between types of citations.
However, papers can be cited for a number of reasons~\citep{bornmann-daniel-2008-citation}, not all positive and not all having to do with the conventions of scholarly publishing \citep{bornmann-daniel-2008-citation,zhu-etal-215-measuring,bornmann-marx-2012-anna}.

\paragraph*{Limitations of this survey}
As with all surveys, our survey results might be subject to selection bias.
To mitigate this risk, we took the following steps:
\textbf{(1)} We used public mailing lists such as corpora-list to advertise our survey outside our personal networks. \textbf{(2)} Our social media and academic networks are diverse as we are authors from four different institutions, covering four different continents, and we are at various career stages (Masters student, PhD candidate, postdoctoral researcher, assistant professor, and full professor). \textbf{(3)} We targeted 100+ survey responses (and received 138).
Despite our efforts to get a large number and diversity of responses, they may not be representative of the field as a whole.
In particular, full professors (N=5, at various career stages), and industry practitioners who are not researchers (N=1) were somewhat underrepresented in our responses, indicating that our results focus more on research impact rather than impact on industry applications, and are mostly shaped by PhD students (41.3\% of respondents), whose interests, incentives, and assessment of impact are sure to be different from respondents at other career stages.

As for survey content, some respondents brought up the following concerns about our design choices: one respondent felt our definition of IA was too broad for their taste, but our inclusion of interpretability \textit{and} analysis was by design (see \Cref{sec:methods}).
Another respondent noted that we defined IA but not what we meant by ``progress,'' which was also by design, as we did not want to impose a normative definition of progress on our respondents but rather, get at their own intuitions, regardless of how they might define progress.
Finally, one respondent complained that our questions about the usefulness of IA (to various subfields, on one's own research, etc.) were framed in absolute rather than relative terms, and that just because IA research has some positive impact on our understanding doesn't mean that it is the best option to pursue given limited time and resources.
This paper presents views of absolute \textit{and} relative impact via the survey and citation graph analyses, for a holistic view of IA research that also allows for it to have value for its own sake. Ultimately, we believe that a view of ``optimal'' impact compared to other options lies in the eye of the beholder, and is one (but not the only) way of interpreting our results.

\section*{Acknowledgments}

We are grateful to Julian Schnitzler, Maor Ivgi, Siva Reddy, Vlad Niculae, Yanai Elazar, and Yonatan Belinkov for their feedback on the survey, as well as Asma Ghandeharioun, Yanai Elazar, and Sabrina Mielke for their feedback on the manuscript.
We would like to thank Anna Rogers, David Chiang, Fei Xia, Henning Wachsmuth, Jordan Lee Boyd-Graber, Juan Pino, Naoaki Okazaki, Rachele Sprugnoli, and Scott Yih, for their help in providing us with ACL and EMNLP track data.
Finally, we thank all our survey respondents, including, among others: AG, AW, Aaron Mueller, Aengus Lynch, Alessandro Stolfo, Alon Jacovi, Anubrata Das, Aryaman Arora, Avi Caciularu, Benjamin Minixhofer, Bhawna Paliwal, Christopher Potts, Chunyuan Deng, Daniel C.H. Tan, Daniel Scalena, Dashiell Stander, David Adelani, David Bau, David Chanin, Diego Garcia-Olano, Emilio Villa-Cueva, Eran Hirsch, Eva Portelance, Felix Beierle, Florian Schneider, Gabriele Sarti, Guanlin Li, Jaap Jumelet, Jack Merullo, Jiahao Huang, Jonathan Zea, Julian Schnitzler, Keshav Ramji, Leshem Choshen, Lucas E. Resck, Margarita Bugueño, Miaoran Zhang, Mircea Petrache, Natalie Shapira, Nils Feldhus, Noah Y. Siegel, Ori Ram, Paulina, Peter Hase, Qinan Yu, Ricardo Cuervo, Roma Patel, Sebastian Breguel, Tian Yun, Tomasz Limisiewicz, Vaidehi Patil, Victor Faraggi, Wentao Wang, Yeo Wei Jie, Yindong Wang, Yonathan Arbel, and Yuval Pinter.

TVB was funded by the Centro Nacional de Inteligencia Artificial, CENIA, FB210017, Basal ANID, MM was supported by the Mila-Samsung grant, and VG was funded by the BMBF’s (German Federal Ministry of
Education and Research) SLIK project under the
grant 01IS22015C.

\bibliography{bibliography}

\appendix

\section{Citation graph details}
\label{sec:appendix:citation-graph}

We provide additional details on the creation of our citation graph below. 

\begin{table}[t]
\centering
\footnotesize %
    \begin{tabularx}{\linewidth}{Xr}
        \toprule
        \textbf{Track} & \textbf{Paper Count} \\
        \midrule
         Information Extraction/Retrieval & 674 \\
         Machine Translation and Multilinguality & 594 \\
         Machine Learning & 557 \\
         Applications & 516 \\
         Dialogue & 487 \\
         Interpretability and Analysis & 477 \\
         Semantics & 456 \\
         Resources and Evaluation & 423 \\
         Multimodality, Speech and Grounding & 389 \\
         Generation & 361 \\
         Question Answering & 334 \\
         Sentiment Analysis & 258 \\
         Summarization & 244 \\
         Theme & 188 \\
         Social Science & 178 \\
         Ethics & 130 \\
         Syntax & 121 \\
         Efficient Methods & 113 \\
         Linguistic Theories and Psycholinguistics & 106 \\
         Discourse and Pragmatics & 84 \\
         Large Language Models & 83 \\
         Industry & 76 \\
         Phonology, Morphology and Word Segmentation & 72 \\
         Commonsense Reasoning & 32 \\
         Human-Centered NLP & 18 \\
         Unsupervised and Weakly Supervised Methods in NLP & 17 \\
         Theory and Formalism in NLP & 6 \\
        \bottomrule
    \end{tabularx}
\caption{Papers per track in ACL/EMNLP.}
\label{tab:research-track-distribution}
\end{table}

\paragraph{Summary statistics}

\Cref{tab:research-track-distribution} shows the number of papers per track in our initial collection. With 477 papers, IA is the 6th largest track in the collection.

\paragraph{Standarizing submission tracks}
The submission tracks of ACL and EMNLP conferences have changed considerably from 2018 to 2023. Some tracks were split into multiple tracks, some tracks appeared (and disappeared), and some were renamed.
As we are mostly interested in comparing IA with other tracks, we decided to merge tracks in order to create a consistent set of tracks starting from 2020 (when the IA track was established). This unification makes our analysis more feasible. We manually assigned every track from ACL/EMNLP from 2020 to 2023 into 27 different categories:
\vspace{0.5em}

\textbullet Information Extraction/Retrieval

\textbullet Machine Translation and Multilinguality

\textbullet Machine Learning

\textbullet Applications

\textbullet Dialogue

\textbullet Semantics

\textbullet Interpretability and Analysis

\textbullet Resources and Evaluation

\textbullet Generation

\textbullet Question Answering

\textbullet Multimodality, Speech and Grounding

\textbullet Summarization

\textbullet Sentiment Analysis

\textbullet Theme

\textbullet Social Science

\textbullet Ethics

\textbullet Linguistic Theories and Psycholinguistics

\textbullet Syntax

\textbullet Efficient Methods

\textbullet Discourse and Pragmatics

\textbullet Large Language Models

\textbullet Phonology, Morphology and Word Segmentation

\textbullet Industry

\textbullet Commonsense Reasoning

\textbullet Human-Centered NLP

\textbullet Unsupervised and Weakly-Supervised Methods in NLP

\textbullet Theory and Formalism in NLP
\vspace{0.5em}

\noindent We note that we consider the EMNLP 2023 track: Language Modeling and Analysis of Language Models as part of IA.
Additionally, we ignore papers from the theme track, as these topics change every year.

\paragraph{Cleaning the collected data}

\begin{table}[t]
\centering
    \resizebox{\columnwidth}{!}{%
    \begin{tabular}{ll}
        \toprule
        \textbf{Statistic} & \textbf{Value} \\
        \midrule
        Nodes (papers)    & 185,384 \\
        Edges (citations)   & 786,376 \\
        Nodes originally from ACL/EMNLP 2018-2023 & 9,248 \\
        References from ACL/EMNLP 2018-2023 papers & 374,857 \\ 
        Citations of ACL/EMNLP 2018-2023 papers & 469,580 \\ 
        \bottomrule
    \end{tabular}%
    }
\caption{Statistics of the citation graph. As some EMNLP/ACL papers cite other EMNLP/ACL papers, the total number of edges is less than the sum of the references and citations.}
\label{tab:citation-graph-statistics}
\end{table}

Since the ACL Anthology does not provide information about the submission track, we obtain our data from a diverse set of sources as listed in \Cref{tab:conference-papers-sources}. Since the data comes in very different formats, we performed the following steps to clean it. 

We searched for paper titles in the ACL anthology to obtain their DOIs. As some papers were renamed, preventing us from finding the corresponding paper in the ACL Anthology, we queried the Semantic Scholar API for the closest match, with a minimum of 0.85 similarity using the Python \verb|difflib.SequenceMatcher| class. Finally, we manually searched for the remaining papers on Semantic Scholar. After this process, we were left with only 6 papers with no Semantic Scholar ID. We exclude these from our analysis.
Finally, for each paper, we queried its citations and its references using the Semantic Scholar API, and constructed the citation graph based on the results.

\paragraph{Citation intent and influence}

For each citation, the Semantic Scholar API provides a label of the intent (e.g. as background information, use of methods, or comparing results) \cite{cohan2019structural}, and a label on whether it is a ``highly influential'' citation for the paper or not \cite{valenzuela2015identifying}. We rely on the latter label when analyzing the most cited IA papers in \Cref{sec:positive-examples}.

\paragraph{Track classifiers details}

We are interested in analyzing how papers from different tracks cite each other. However, as most of the nodes in our citation graph are papers that are not in ACL and EMNLP, we have no ground truth information for the track of these papers. Therefore, we built a classifier to predict the track of a paper, given its title and abstract. The classifier is based on the Specter2 model \cite{cohan2020specter}, which takes a title and an abstract of a paper, and outputs an embedding. We add and train a MLP layer on top of this model to obtain our classifier.

We split the data 80/20 using only papers from ACL and EMNLP from 2020 to 2023 (for which we have gold labels), and we trained the classifier for 50 epochs using Adam and a cross entropy loss. We used a learning rate of $2 * 10^{-3}$ and a learning rate scheduler with exponential decay ($\gamma = 0.995$). We perform upsampling as the number of papers in each track is imbalanced. Additionally, to get an even more diverse set of papers for the interpretability and analysis track, we augment the training data with papers accepted to the BlackboxNLP workshop, which focuses on IA work.

\begin{table}[t]
\centering
    \resizebox{\columnwidth}{!}{%
    \begin{tabular}{ll}
        \toprule
        \textbf{Conference} & \textbf{Data Source} \\
        \midrule
        ACL 2018 & Conference schedule web page \\
        ACL 2019 & Conference schedule web page \\
        ACL 2020 & Virtual conference web page \\
        ACL 2021 & Conference schedule web page \\
        ACL 2022 & Provided by the program chairs \\
        ACL 2023 & Github repository to generate webpage \\
        EMNLP 2018 & Provided by the program chairs \\
        EMNLP 2019 & Conference schedule web page \\
        EMNLP 2020 & Github repository to generate webpage \\
        EMNLP 2021 & Provided by the program chairs \\
        EMNLP 2022 & Provided by the program chairs \\
        EMNLP 2023 & Provided by the program chairs \\
        \bottomrule
    \end{tabular}%
    }
\caption{Data source for each conference.}
\label{tab:conference-papers-sources}
\end{table}

We find that some tracks are more difficult to predict correctly than others (e.g., Efficient Methods). We attribute this to both the limited training data and the ambiguity of submission tracks. We hence restrict ourselves to the 11 tracks (including IA) with the highest classification accuracy, and introduced an `Other' category to group the remaining tracks, which we exclude from our classifier analyses. The final set of tracks in our classifier is:

\textbullet    Dialogue

\textbullet    Ethics

\textbullet    Generation

\textbullet    Information Extraction/Retrieval

\textbullet    Interpretability and Analysis

\textbullet    Machine Learning

\textbullet    Machine Translation and Multilinguality

\textbullet    Multimodality, Speech and Grounding

\textbullet    Question Answering

\textbullet    Social Science

\textbullet    Summarization

\textbullet    Other

On this final set of tracks, our classifier achieves an F1 micro/macro score of 
0.61/0.61.
Given how noisy submission track labels can be (a paper can often be a plausible candidate for multiple tracks), we find our classifier's performance to be reasonable. We additionally perform a manual error analysis and expect the classification errors made on the test set; most errors were cases where the paper could have been submitted to the predicted track.

Finally, we label the citation graph using our classifier. We used Semantic Scholar and OpenAlex \cite{priem2022openalex} (in accordance with their terms of use) to obtain abstracts. 4.9\% of the papers had no abstract in either source; we thus exclude these from our analysis.

\subsection{Sanity checks}

\paragraph{Additional IA track classifier evaluations}
\label{sec:classifier-evals}

As we are mostly interested in the performance of detecting IA papers, we validate our classifier in 2 different ways: using the IA papers suggested by our respondents in the survey, and manual annotation of 556 papers.

For papers suggested by survey respondents (after removing papers included in the training data), 
we run our classifier and get predicted tracks. The classifier obtained an accuracy of 78.1\% (82/105).
Considering that these papers are out-of-domain in comparison to the training data (some are even IA papers outside of NLP), we believe this to be a good result.

As for the 556 papers that were manually annotated by two authors, our classifier is 87.8\% (488/556) accurate.
As this data is biased towards non-IA papers (506/556 papers), we also compute precision, recall and F1 scores.
The F1 score is 0.60, precision is 1.0 and recall is 0.42.
Since high precision and low recall show that we underselect IA papers, we get a conservative estimate of our positive results rather than an overly generous estimate, which we find acceptable.

\paragraph{Citation trends of IA exclusively inside ACL and EMNLP}
\label{sec:citation-inside-acl-emnlp}

As some of our findings depend on labels from our classifier, which might be noisy, we verify these findings using a smaller subset of our data, consisting exclusively of ACL and EMNLP papers.
This subgraph of our citation graph has gold labels for the submission track.
Specifically, we verify that \textbf{(1)} IA work is primarily cited by tracks other than IA (see \Cref{fig:interp-vs-non-interp}), and that \textbf{(2)} there is significant variation in how frequently different tracks cite IA work (see \Cref{fig:tracks-citing-interp}).\looseness=-1

In our gold labeled subgraph, there are 2,283 citations to IA papers, of which only 846 are from other IA papers (37.1\%).
This shows that a large fraction of citations to IA papers do indeed come from outside IA research.
This verifies our first classifier-based result.

Next, when looking at the references of papers in each submission track within our gold subgraph, we find that the proportion of references to IA papers does indeed differ considerably by track.
As an example, for the Large Language Models track, we find that 11.2\% (N=723) of its references are to IA papers.
In contrast, only 1.3\% (N=1828) of the references of Sentiment Analysis track papers correspond to IA work.
This confirms our second classifier-based finding.

\paragraph{Correlation between betweenness centralities and citation counts}
\label{sec:normalizations}

\begin{figure}[t]
    \centering
    \includegraphics[width=0.85\columnwidth]{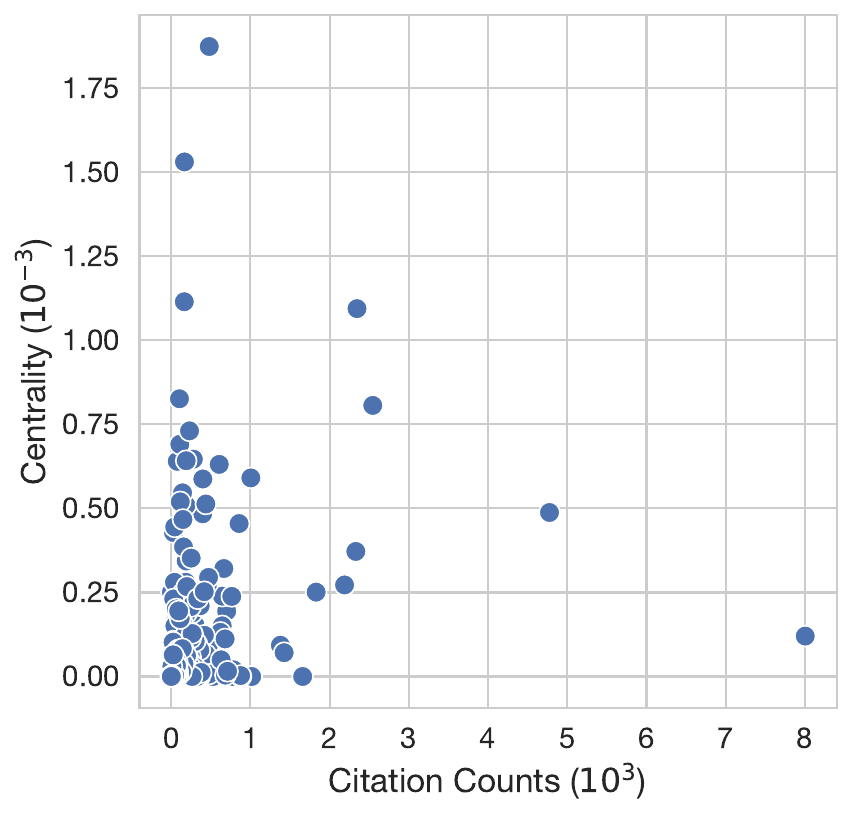}
        \caption{Betweenness centralities versus citation counts for papers in ACL and EMNLP since 2020. No correlation can be detected to the naked eye between these metrics.}
    \label{fig:centrality-vs-citations}
\end{figure}

\citet{leydesdorff2007betweenness} find that betweenness centrality can be highly correlated to citation counts. Although this is expected (papers with more citations can also act better as \textit{bridges}), given that BC is being used as a proxy to measure the ``interdisciplinarity" of a field, we would want this metric to be somewhat orthogonal to the citation counts. We compute the the correlation between the citation counts and the BC of all nodes in our citation graph. At 0.328 ($p < 0.001$), it is considerably lower than the 0.509 reported by \citet{leydesdorff2007betweenness}. \Cref{fig:centrality-vs-citations} provides a visualization of the correlation.

\section{Survey details}
\label{sec:appendix:survey}

We outline ethical considerations pertaining to our survey, along with the final version of the survey below.

\subsection{Ethical considerations}

Our survey involved research with human participants, thus we report the full text of the survey below, and information about recruitment in \Cref{sec:methods}.
We determined there to be a negligible risk of harms from participating in our survey, as it contains no offensive or harmful content.
As shown in the full survey below, we describe our study objectives and remind respondents that filling out the survey is completely voluntary.
We then explicitly ask for their consent to participate, and obtain consent from all 138 survey respondents.
For respondents who may not have completed the survey, no data was collected. In lieu of financial compensation, we offered survey respondents the optional opportunity to provide their name or an alias that we would mention in the acknowledgements of any future paper we write with the survey results.
To protect respondent privacy and confidentiality, when we report disaggregated results in this paper, we ensured a minimum of 10 respondents per bucket.
In addition, we will not release the original survey responses in full, but only release high-level statistics, annotations from our qualitative coding, and select non-identifying examples in \Cref{sec:going-forward}.

\subsection{Participant demographics}

We collected demographic information (occupations and research areas) from survey respondents to consider factors that might affect the representativeness of our results.
\Cref{tab:participant-demographics-occupation} presents the breakdown of respondents per occupation.

\begin{table}
    \centering
    \begin{tabularx}{\linewidth}{Xr}
        \toprule
        \textbf{Occupation} & \textbf{Responses} \\
        \midrule
        PhD student/candidate & 57 (\phantom{0}41\%) \\
        Postdoc & 15 (\phantom{0}11\%) \\
        Junior industry researcher & 17 (\phantom{0}12\%) \\
        Master's student & 12 (\phantom{00}8\%) \\
        Assistant professor & 10 (\phantom{00}7\%) \\
        Senior industry researcher & 10 (\phantom{00}7\%) \\
        Bachelor's student & 6 (\phantom{00}5\%) \\
        Full professor & 5 (\phantom{00}4\%) \\
        Associate professor & 2 (\phantom{00}1\%) \\
        NLP Practitioner & 1 (\phantom{00}1\%) \\
        Other (write-in) & 3 (\phantom{00}2\%) \\
        \midrule
        Total & 138 (100\%) \\
         \bottomrule
    \end{tabularx}
    \caption{Raw numbers and percentages of survey responses, grouped by the respondent's occupation.}
    \label{tab:participant-demographics-occupation}
\end{table}

When collecting information on research areas, we allowed respondents to check multiple boxes corresponding to multiple research areas.
Of particular interest is the area labeled "Science of LMs," which we used as an umbrella term to include analysis and interpretability research.
\Cref{tab:participant-demographics-research-area} shows the research areas of our respondents.
The next section provides the expansions for each of the umbrella terms that we list in the table.

\begin{table}
    \centering
    \begin{tabular}{lr}
        \toprule
        \textbf{Research area} & \textbf{Responses} \\
        \midrule
        Science of LMs & 54 (39\%) \\
        Evaluation & 53 (38\%) \\
        LM adaptation & 47 (34\%) \\
        Data for LMs & 32 (23\%) \\
        NLP applications & 32 (23\%) \\
        Computational linguistics & 30 (22\%) \\
        Mind, brain and LMs & 30 (22\%) \\
        Neurosymbolic approaches & 26 (19\%) \\
        Learning algorithms & 25 (18\%) \\
        LMs for everyone & 24 (17\%) \\
        LMs and the world & 21 (15\%) \\
        Safety & 21 (15\%) \\
        Societal implications & 20 (14\%) \\
        Inference algorithms & 14 (10\%) \\
        Multimodal and novel applications & 14 (10\%) \\
        Compute-efficient LMs & 10 (\phantom{0}7\%) \\
         \bottomrule
    \end{tabular}
    \caption{Raw numbers and percentages of survey respondents who selected a certain research area. Respondents were allowed to select multiple areas, which is why the numbers add up to more than 138. Refer to the full survey for details on what each umbrella term represents.}
    \label{tab:participant-demographics-research-area}
\end{table}

\subsection{Full survey}

\subsection*{Impact of Model Analysis and Interpretability Research on Progress in NLP}

\noindent Estimated time to complete the survey: 12 minutes

\subsubsection*{Study description}

This project aims to measure the impact that model analysis and interpretability research has on current progress in NLP as well as its possible future impact on the field. \\

\noindent You are encouraged to fill out this survey even if you have no exposure to model analysis and interpretability work. \\

\noindent Filling out this questionnaire is completely voluntary. \\

\noindent By clicking "Yes" below, I am verifying that I have read the description above and I consent to participate in this research study.

\noindent \textbullet \ Yes \\
\noindent \textbullet \ No

\subsubsection*{What do we mean by model analysis and interpretability research?}

\noindent Model analysis and interpretability research in natural language processing (NLP) aims to develop a deeper understanding of and explain the behavior of NLP systems. \\

\noindent This includes (but is not limited to) explaining models’ internal computations, investigating broader phenomena observed during pre-training or adaptation, and providing a better understanding of the limitations and robustness of existing models. \\

\noindent Work on topics such as attribution methods, probing, mechanistic interpretability, analysis of embedding spaces, explainability, analysis of training dynamics, analyzing model bias, etc., are additional examples of model analysis and interpretability research.

\subsubsection*{Background questions}

\noindent \textbf{1. What is your occupation?}

\noindent \textbullet \ Bachelor's student \\
\noindent \textbullet \ Master's student \\
\noindent \textbullet \ PhD student/candidate \\
\noindent \textbullet \ Postdoc \\
\noindent \textbullet \ Assistant professor \\
\noindent \textbullet \ Associate professor \\
\noindent \textbullet \ Full professor \\
\noindent \textbullet \ Junior industry researcher \\
\noindent \textbullet \ Senior industry researcher \\
\noindent \textbullet \ NLP practitioner \\
\noindent \textbullet \ Other [fill in] \\

\noindent \textbf{2. What is your area of research?}

\noindent Feel free to select multiple options or add missing ones. 

\noindent \textit{(The list below is adapted from the calls for papers of COLM and ARR.)}

\noindent \textbullet \ \textbf{LM adaptation}: fine-tuning, instruction-tuning, reinforcement learning (with human feedback), prompt tuning, and in-context alignment \\
\noindent \textbullet \ \textbf{Data for LMs}: pre-training data, alignment data, and synthetic data --- via manual or algorithmic analysis, curation, and generation \\
\noindent \textbullet \ \textbf{Evaluation of LMs}: benchmarks, simulation environments, scalable oversight, evaluation protocols and metrics, human and/or machine evaluation \\
\noindent \textbullet \ \textbf{Societal implications}: bias, fairness, accountability, transparency, equity, misuse, jobs, climate change, and beyond \\
\noindent \textbullet \ \textbf{Safety}: security, privacy, misinformation, adversarial attacks and defenses \\
\noindent \textbullet \ \textbf{Science of LMs}: scaling laws, fundamental limitations, emergent capabilities, demystification, interpretability, complexity, training dynamics, grokking, learning theory for LMs \\
\noindent \textbullet \ \textbf{Compute efficient LMs}: distillation, compression, quantization, sample efficient methods, memory efficient methods \\
\noindent \textbullet \ \textbf{Engineering for large LMs}: distributed training and inference on different hardware setups, training dynamics, optimization instability \\
\noindent \textbullet \ \textbf{Learning algorithms}: learning, unlearning, meta learning, model mixing methods, continual learning \\
\noindent \textbullet \ \textbf{Inference algorithms}: decoding algorithms, reasoning algorithms, search algorithms, planning algorithms \\
\noindent \textbullet \ \textbf{Human mind, brain, philosophy, laws and LMs}: cognitive science, neuroscience, linguistics, psycholinguistics, philosophical, or legal perspectives on LMs \\
\noindent \textbullet \ \textbf{LMs for everyone}: multilinguality, low-resource languages, vernacular languages, multiculturalism, value pluralism \\
\noindent \textbullet \ \textbf{LMs and the world}: factuality, retrieval-augmented LMs, knowledge models, commonsense reasoning, theory of mind, social norms, pragmatics, and world models \\
\noindent \textbullet \ \textbf{LMs and embodiment}: perception, action, robotics, and multimodality \\
\noindent \textbullet \ \textbf{LMs and interaction}: conversation, interactive learning, and multi-agents learning \\
\noindent \textbullet \ \textbf{LMs with tools and code}: integration with tools and APIs, LM-driven software engineering \\
\noindent \textbullet \ \textbf{LMs on diverse modalities and novel applications}: visual LMs, code LMs, math LMs, and so forth, with extra encouragements for less studied modalities or applications such as chemistry, medicine, education, database and beyond \\
\noindent \textbullet \ \textbf{NLP applications}: sentiment analysis, summarization, question answering, etc. \\
\noindent \textbullet \ \textbf{Computational linguistics}: discourse, pragmatics, phonology, morphology, syntax, semantics \\
\noindent \textbullet \ \textbf{Information extraction, information retrieval, text mining} \\
\noindent \textbullet \ \textbf{Neurosymbolic approaches} \\
\noindent \textbullet \ \textbf{Non-neural methods approaches for NLP} \\
\noindent \textbullet \ Other [fill in] \\

\noindent \textbf{[OPTIONAL]}

\noindent \textbf{If you would like, provide your name (or an alias) here and we will mention it in the acknowledgements of our future paper.} [fill in]

\subsubsection*{Your take on model analysis and interpretability research}

\noindent \textbf{Reminder: What do we mean by model analysis and interpretability research?}

\noindent Model analysis and interpretability research in natural language processing (NLP) aims to develop a deeper understanding of and explain the behavior of NLP systems. \\

\noindent This includes (but is not limited to) explaining models’ internal computations, investigating broader phenomena observed during pre-training or adaptation, and providing a better understanding of the limitations and robustness of existing models. \\

\noindent Work on topics such as attribution methods, probing, mechanistic interpretability, analysis of embedding spaces, explainability, analysis of training dynamics, analyzing model bias, etc., are additional examples of model analysis and interpretability research. \\

\noindent \textbf{3. How much do you agree with the following statement?}

\noindent The progress in NLP in the last five years would \textbf{not have been possible} without findings from model analysis and interpretability research.

\noindent \textbullet \ 1: strongly disagree \\
\noindent \textbullet \ 2 \\
\noindent \textbullet \ 3 \\
\noindent \textbullet \ 4 \\
\noindent \textbullet \ 5: strongly agree \\

\noindent \textbf{4. How much do you agree with the following statement?}

\noindent The progress in NLP in the last five years would have been \textbf{slower} without findings from model analysis and interpretability research.

\noindent \textbullet \ 1: strongly disagree \\
\noindent \textbullet \ 2 \\
\noindent \textbullet \ 3 \\
\noindent \textbullet \ 4 \\
\noindent \textbullet \ 5: strongly agree \\

\noindent \textbf{5. How many model analysis and interpretability works do you read compared to other topics?}

\noindent \textbullet \ I don't usually read model analysis and interpretability work, but I do read NLP works about other topics \\
\noindent \textbullet \ I do read some model analysis and interpretability work, but much less than other topics \\
\noindent \textbullet \ I read model analysis and interpretability work in about the same volume as other NLP-related topics \\
\noindent \textbullet \ I read model analysis and interpretability work more than other NLP topics \\
\noindent \textbullet \ Most of the works I read are about model analysis and interpretability \\

\noindent \textbf{6. How, if at all, does model analysis and interpretability work influence your own work?}

\noindent $\square$ It provides me with new research ideas \\
\noindent $\square$ It changes my mental model of what the capabilities and limitations of models are \\
\noindent $\square$ It helps me ground my explanations of my own results \\
\noindent $\square$ It adds useful tools for me to visualize/evaluate/understand the behavior of a model \\
\noindent $\square$ It does not influence my work \\
\noindent $\square$ Other [fill in] \\

\noindent \textbf{[OPTIONAL]}

\noindent \textbf{7. Provide up to 5 model analysis and interpretability papers that have influenced your work (please provide a comma separated list of paper titles or URLs).} [fill in] \\

\noindent \textbf{8. In your day-to-day work, do you use concepts from model analysis and interpretability research (e.g., probing, residual stream, induction heads, causal interventions, MLP layers as key-value memories, etc.)?}

\noindent \textbullet \ Never \\
\noindent \textbullet \ Rarely \\
\noindent \textbullet \ Sometimes \\
\noindent \textbullet \ Often \\
\noindent \textbullet \ Always \\

\noindent \textbf{9. Do you think model analysis and interpretability research is important, and if so, why?}

\noindent $\square$ Understanding model limitations and capabilities \\
\noindent $\square$ Making models more computationally efficient \\
\noindent $\square$ Developing safety mechanisms \\
\noindent $\square$ Improving model trustworthiness \\
\noindent $\square$ Explainability for users \\
\noindent $\square$ To fullfill legal requirements (e.g., GDPR) \\
\noindent $\square$ Improving model capabilities \\
\noindent $\square$ Developing novel architectures \\
\noindent $\square$ Developing novel architectures \\
\noindent $\square$ I do not think model analysis and interpretability work is important \\
\noindent $\square$ Other [fill in] \\

\noindent \textbf{[OPTIONAL]}

\noindent \textbf{10. If you selected "I do not think model analysis and interpretability research is important" above, please elaborate why.} [fill in] \\

\noindent \textbf{[OPTIONAL]}

\noindent \textbf{11. In your opinion, how important is model analysis and interpretability research to work in the areas below?} \\

\noindent Work on multilinguality and low-resource languages \\
\noindent \textbullet \ Model analysis and interpretability research is not important for \\
\noindent \textbullet \ Model analysis and interpretability research is somewhat important for \\
\noindent \textbullet \ Model analysis and interpretability research is very important for \\

\noindent Work on multimodal learning, grounding, and embodiment \\
\noindent \textbullet \ Model analysis and interpretability research is not important for \\
\noindent \textbullet \ Model analysis and interpretability research is somewhat important for \\
\noindent \textbullet \ Model analysis and interpretability research is very important for \\

\noindent Work on engineering for large language models \\
\noindent \textbullet \ Model analysis and interpretability research is not important for \\
\noindent \textbullet \ Model analysis and interpretability research is somewhat important for \\
\noindent \textbullet \ Model analysis and interpretability research is very important for \\

\noindent Work on factuality, reasoning, world models \\
\noindent \textbullet \ Model analysis and interpretability research is not important for \\
\noindent \textbullet \ Model analysis and interpretability research is somewhat important for \\
\noindent \textbullet \ Model analysis and interpretability research is very important for \\

\noindent Work on societal implications, bias, misuse, and beyond \\
\noindent \textbullet \ Model analysis and interpretability research is not important for \\
\noindent \textbullet \ Model analysis and interpretability research is somewhat important for \\
\noindent \textbullet \ Model analysis and interpretability research is very important for \\

\noindent \textbf{[OPTIONAL]}

\noindent \textbf{12. In your opinion, what is missing in model analysis and interpretability research right now? Where should it go in the future and how should it be shaped differently?} [fill in] \\

\noindent \textbf{[OPTIONAL]}

\noindent \textbf{13. Do you have additional opinions or thoughts on model analysis and interpretability research?} [fill in]

\section{Qualitative coding}
\label{sec:qualitative-coding}

Qualitative coding is an inductive methodology from the social sciences~\citep{Saldana2021-ki}, used to systematically surface thematic patterns in data with less structure
In the context of this paper, we use qualitative coding to analyze open-ended survey responses, and paper titles and abstracts.
Two authors performed qualitative analysis of all 70 open-ended survey responses, and 556 papers (based on their titles and abstracts).

We began by analyzing the survey responses:
one round of independent coding was done, based on which we reviewed our codes to normalize terms and resolve disagreements.
After this, a second round of annotation was performed.

As for the paper annotations, the authors
did a combination of independent coding (with discussion and re-coding), and co-coding.
Throughout the annotation process, the authors followed best practices by working closely together to clarify the annotation procedure, discuss the emerging themes, and re-annotate data that was coded early on~\citep{BENGTSSON20168}.

We iteratively merged codes for related themes (e.g., \textit{pre-training trajectories} and \textit{training dynamics}), and to resolve inconsistencies from typos (e.g., \textit{in-context learning} instead of \textit{in-contex learning}) and to normalize themes (e.g., \textit{interventions} instead of \textit{intervention}), where applicable.
All merging operations are released as part of our code.

We measure inter-coder reliability with percentage agreement~\citep{oconnorjoffe2020icr}, which was above 90\% across all subsets of annotation. Summary statistics are shown in Table \ref{tab:qualitative-coding}.

\begin{table*}
    \centering
    \begin{tabularx}{\linewidth}{Xrrrr}
        \toprule
        \textbf{Data source} & \textbf{Instances} & \textbf{Themes (total)} & \textbf{Themes (per instance)} & \textbf{Agreement} \\
        \midrule
        Survey (what's missing?) & 42 & 44 & 2.12 & 91.01 \\
        Survey (why not important?) & 6 & 9 & 1.5 & 100.00 \\
        Survey (additional thoughts) & 22 & 29 & 1.95 & 100.00 \\
        Papers (survey) & 29 & 59 & 4.28 & 100.00 \\
        Papers (top-50 IA) & 50 & 115 & 5.38 & 97.03 \\
        Papers (top-50 non-IA) & 50 & 99 & 4.46 & 96.41 \\
        Papers (non-IA papers highly influenced by IA) & 456 & 327 & 4.90 & 97.49 \\
         \bottomrule
    \end{tabularx}
    \caption{Qualitative coding statistics. For each data source, we list the total number of data instances, the total number of themes assigned, the number of themes per instance, and the percentage agreement between the codes assigned by two annotators.}
    \label{tab:qualitative-coding}
\end{table*}

\section{Additional results}
\label{sec:appendix:additional-results}

\paragraph{Relative growth of submission tracks}

\begin{figure}[t]
    \centering
    \includegraphics[width=\linewidth]{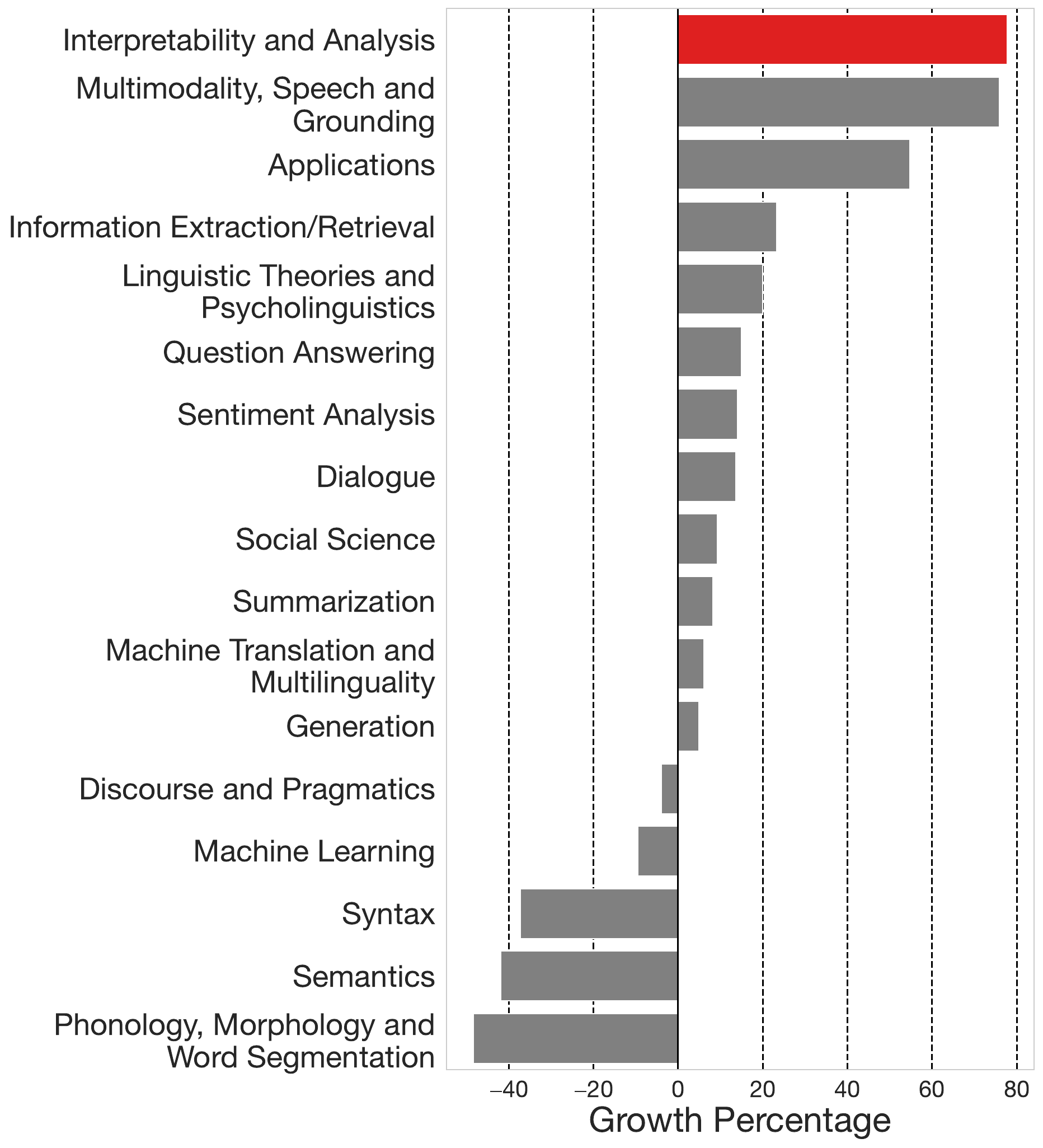}
        \caption{Growth of accepted papers per track in comparing ACL/EMNLP in 2020 vs. in 2023. This considers the tracks that have consistently existed in ACL and EMNLP in both those years.}
    \label{fig:growth-per-track}
\end{figure}

\Cref{fig:growth-per-track} shows the the relative growth of the IA track compared to other tracks that have consistently existed since 2020. IA is the fastest growing track at ACL and EMNLP.

\paragraph{Betweenness centrality}

\begin{figure*}[t]
    \centering
    \includegraphics[width=\linewidth]{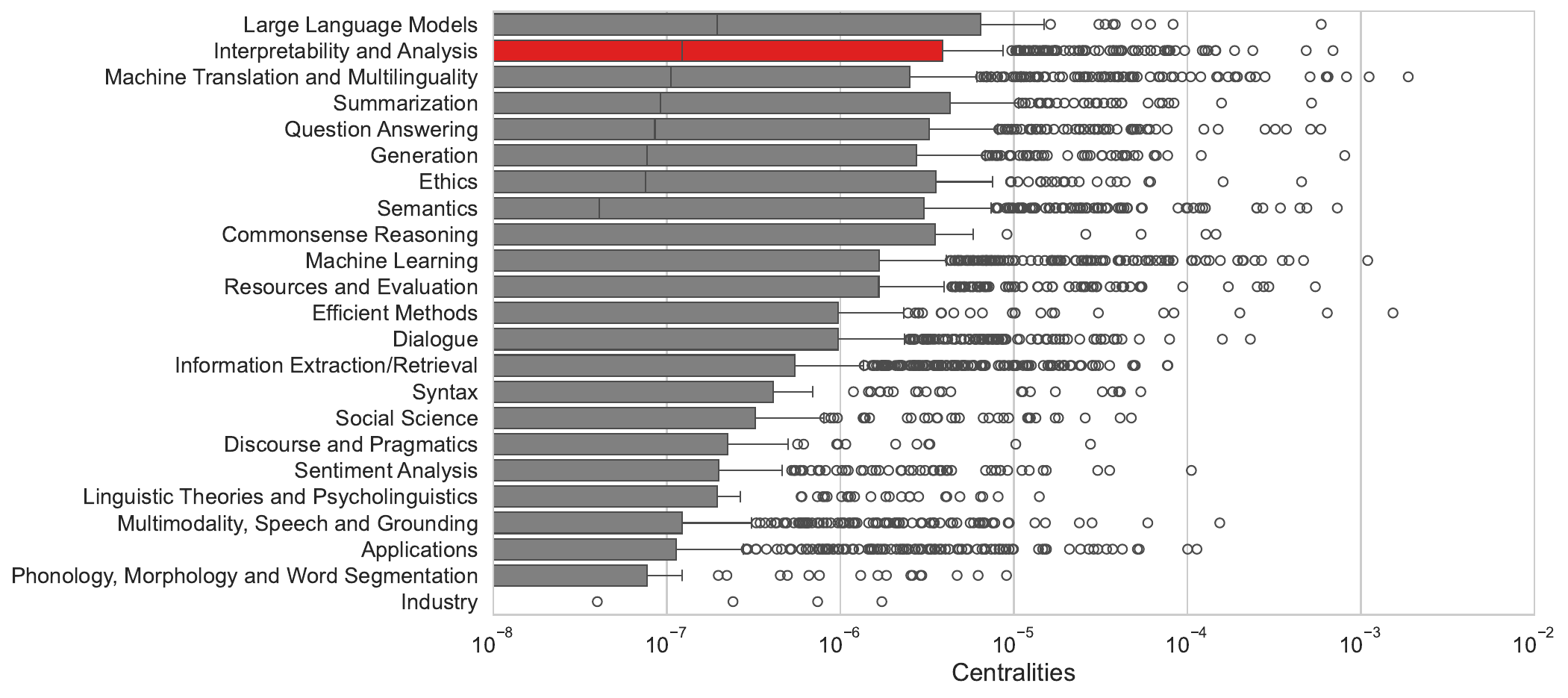}
        \caption{Betweenness centrality of ACL and EMNLP papers since 2020 by track. Lines at the middle of the box represent the medians, but some tracks have their median at 0. IA papers are more \textit{central} than papers from most tracks.\looseness=-1
        }
    \label{fig:betweenness-centrality}
\end{figure*}

\Cref{fig:betweenness-centrality} shows the betweenness centralities for the different tracks we consider. We note that for this analysis we only consider the portion of the citation graph for which we have gold track labels. Our results show that IA has the second largest median centrality. This indicates that IA plays a central role in the ACL/EMNLP citation graph, in the sense that IA papers often lie on the shortest path that connects to random papers of the graph. 

\paragraph{Which tracks cite IA papers}

\begin{figure*}[t]
    \centering
     \includegraphics[width=\linewidth]{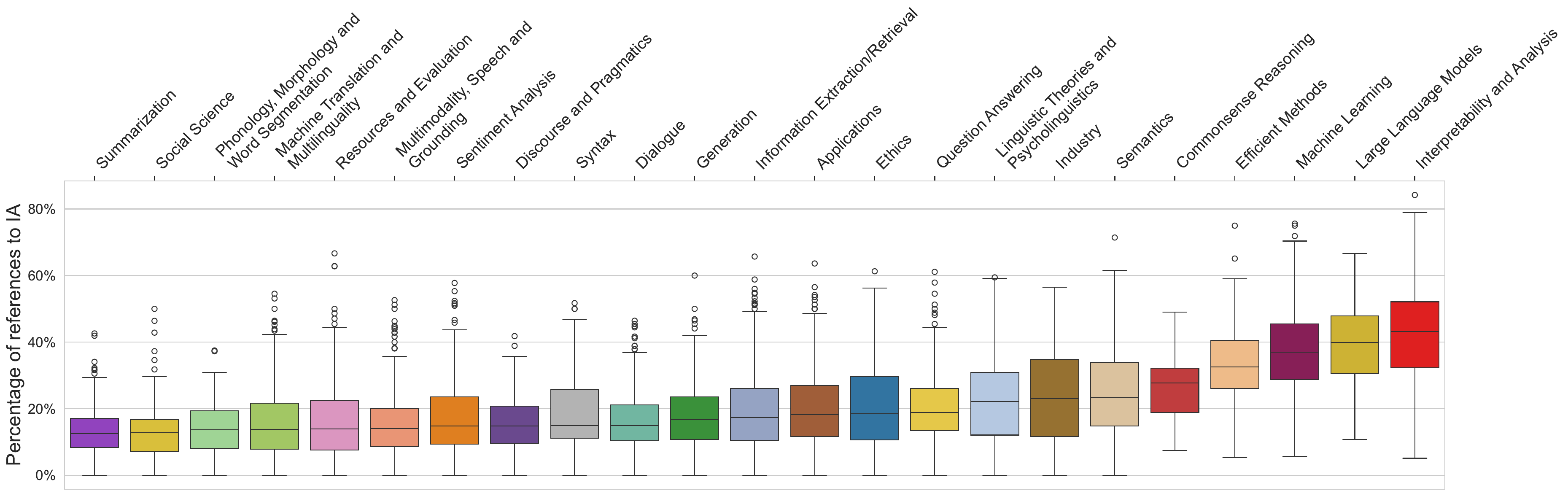}
        \caption{Percentage of references to IA papers according to our classifiers prediction for different tracks. There are significant differences across tracks in how IA is cited. This is also true when only considering gold labels for tracks (see \Cref{sec:citation-inside-acl-emnlp}).}
    \label{fig:tracks-citing-interp}
\end{figure*}

\begin{figure}[t]
    \centering        \includegraphics[width=0.48\textwidth]{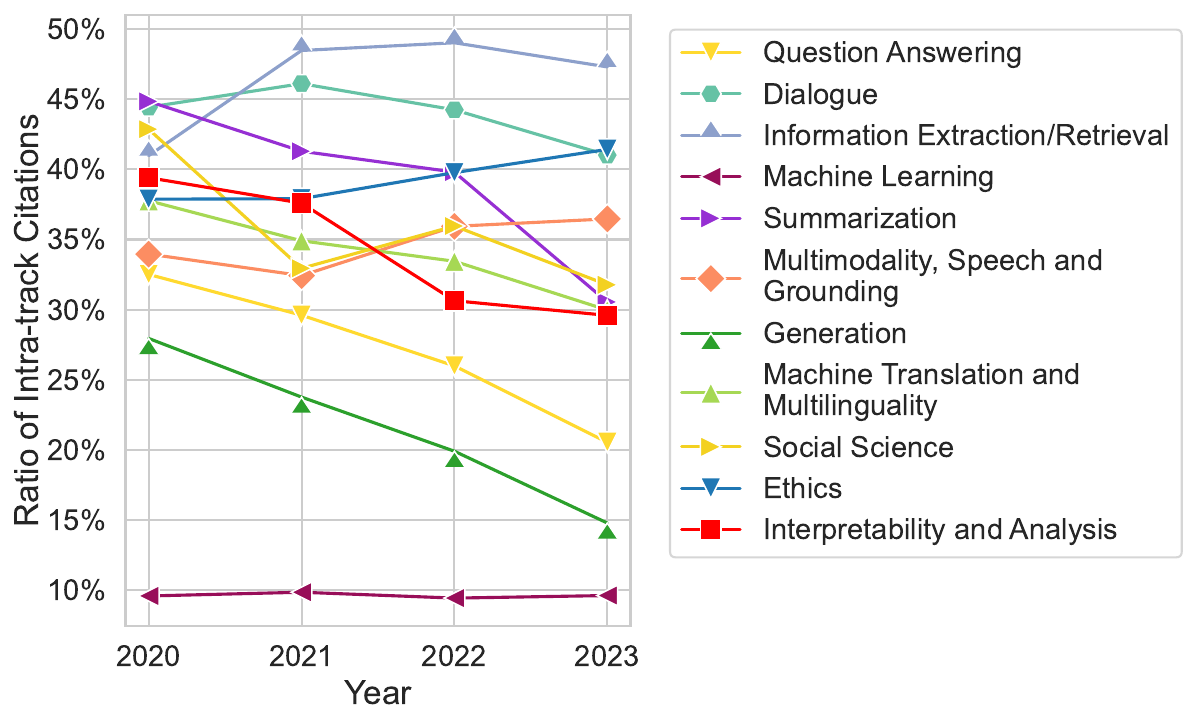}
        \caption{Ratio of intra-track citations according to the predictions of our classifier. It measures the percentage of citations to papers of track A from papers that are also in track A. IA does not stand out in terms of the percentage of citations which are made by other papers of its own track.}
    \label{fig:intra-track}
\end{figure}

\Cref{fig:tracks-citing-interp} shows the percentage of references to IA papers across tracks. Efficient Methods, Machine Learning, and Large Language Models cite IA papers more often than other tracks. 

\paragraph{Comparing extra-track ratios}

\Cref{fig:intra-track} compares the percentage of intra-track citations across tracks. The percentage of intra-track citations of the IA track is positioned roughly in the middle of tracks. This shows that IA is not an outlier in terms of intra-track citations. 

\paragraph{Top themes of highly cited IA papers}

\begin{table*}[t]
    \centering
    \begin{tabularx}{\linewidth}{lX}
    \toprule
    \textbf{Source} & \textbf{Top themes (\% of papers in which the theme appears)} \\
    \midrule
        Survey & representation analysis (34\%), novel method (24\%), probing (24\%), attention analysis (21\%), interventions (17.2\%), mechanistic interp (17.2\%), attribution (17.2\%) \\
        Top-50 IA & analysis (40\%), novel method (36\%), evaluation (32\%), explainability (20\%), linguistics (16\%), probing (16\%) \\
        Top-50 non-IA & novel model (34\%), novel method (32\%), novel dataset (24\%), analysis (16\%) \\
        \bottomrule
    \end{tabularx}
    \caption{Top themes of highly influential IA papers (mentioned by survey respondents and top-50 most-cited IA papers from the citation graph), compared to the top themes of the top-50 most-cited non-IA papers. Themes are not mutually exclusive.}
    \label{tab:influential-paper-themes}
\end{table*}

\Cref{tab:influential-paper-themes} shows the top themes that appear in (1) the papers mentioned by survey participants; (2) the top-50 most cited IA papers; (3) the top-50 most cited non-IA papers.

\paragraph{Citational intent}

\begin{figure*}[t]
    \centering
        \includegraphics[width=\textwidth]{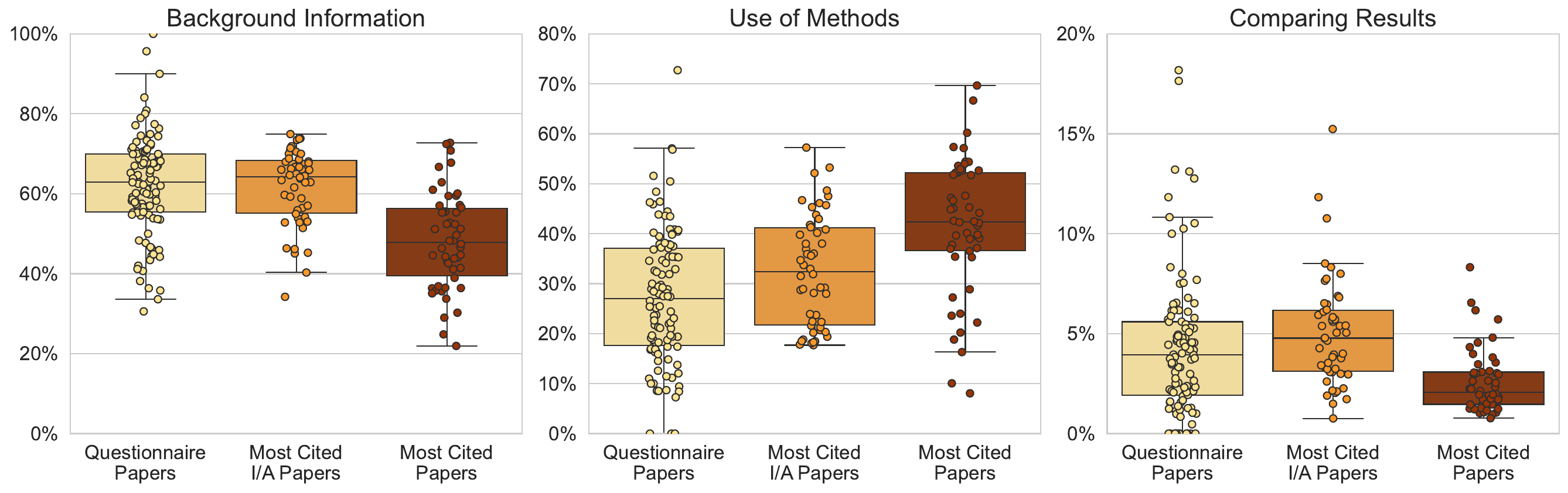}
        \caption{Citation intent percentages for the interpretability and analysis papers suggested in the responses in our survey, the top cited interpretability and analysis papers in ACL/EMNLP, and the top cited papers in ACL/EMNLP for any track.}
    \label{fig:citation-intent-top-papers}
\end{figure*}

\Cref{fig:citation-intent-top-papers} shows the distribution of citation intents for three groups: IA papers suggested in our survey responses, the top cited IA papers in ACL/EMNLP, and the overall most cited papers in ACL/EMNLP within our citation graph. Both the IA papers suggested in our survey and the top cited IA papers in ACL/EMNLP are primarily cited as \textit{background information}. In contrast, the overall top cited papers in ACL/EMNLP are mostly cited for their \textit{use of methods}.

\end{document}